\documentclass[11pt]{article}

\usepackage[preprint]{acl}
\usepackage{multirow}
\usepackage{graphicx}
\usepackage[table,xcdraw]{xcolor}
\usepackage[normalem]{ulem}
\usepackage{float}
\useunder{\uline}{\ul}{}
\usepackage{times}
\usepackage{latexsym}
\usepackage{amssymb}
\usepackage{url}
\usepackage{hyperref}
\usepackage{url}
\usepackage{hyperref}
\usepackage{fontawesome5}
\usepackage[dvipsnames]{xcolor}

\usepackage[T1]{fontenc}

\usepackage[utf8]{inputenc}

\usepackage{microtype}

\usepackage{inconsolata}

\usepackage{booktabs}
\usepackage{comment}

\usepackage{adjustbox}
\usepackage{rotating}

\usepackage[most]{tcolorbox}
\usepackage{listings}
\usepackage{wrapfig,lipsum,booktabs}

\definecolor{codebg}{RGB}{245,245,245}
\definecolor{codeframe}{RGB}{220,220,220}
\definecolor{codetitle}{RGB}{80,80,80}
\definecolor{keycolor}{RGB}{0,0,255}
\definecolor{stringcolor}{RGB}{0,128,0}
\definecolor{commentcolor}{RGB}{128,128,128}

\lstdefinelanguage{yaml}{
  keywords={true,false,null,y,n},
  keywordstyle=\color{keycolor}\bfseries,
  basicstyle=\ttfamily\footnotesize,
  breaklines=true,
  comment=[l]{\#},
  commentstyle=\color{commentcolor}\itshape,
  stringstyle=\color{stringcolor},
  moredelim=[l][\color{orange}]{\&},
  moredelim=[l][\color{orange}]{\*},
  morestring=[b]',
  morestring=[b]",
  literate={-}{{{\color{red}-}}}1 {:}{{{\color{red}:}}}1,
}

\newtcolorbox{codeblock}[2][]{
    enhanced,
    boxrule=0.5pt,
    colback=codebg,
    colframe=codeframe,
    coltitle=codetitle,
    fonttitle=\bfseries\ttfamily\footnotesize,
    title={#2},
    attach boxed title to top left={yshift=-2mm, xshift=2mm},
    boxed title style={colback=white, boxrule=0.5pt, colframe=codeframe},
    arc=2pt,
    breakable,
    #1
}

\usepackage{siunitx}
\usepackage{xfp}
\usepackage{graphicx}

\usepackage{xcolor}
\usepackage{enumitem}
\usepackage{xspace}

\newcommand*{\sym}{\texttt{SYM}\xspace}
\newcommand*{\symn}{\texttt{SYM\_N}\xspace}
\newcommand*{\symhash}{\texttt{SYM\_\#}\xspace}
\newcommand*{\symboth}{\texttt{SYM\_N\#}\xspace}
\newcommand*{\ic}{\texttt{IC}\xspace}
\newcommand*{\icn}{\texttt{IC\_N}\xspace}
\newcommand*{\ichash}{\texttt{IC\_\#}\xspace}
\newcommand*{\icboth}{\texttt{IC\_N\#}\xspace}

\newcommand{\rkup}{\textcolor{ForestGreen}{\faCaretUp}}
\newcommand{\rkdn}{\textcolor{BrickRed}{\faCaretDown}}
\newcommand{\rkno}{\textendash}

\usepackage[table]{xcolor}
\usepackage{colortbl}

\definecolor{heatred}{RGB}{218, 92, 92}

\def\dobase{100}

\newcommand{\dr}[1]{\gdef\dobase{#1}#1}

\newcommand{\hc}[1]{%
  \dimen255=\dimexpr\dobase pt - #1 pt\relax
  \ifdim\dimen255<1pt #1%
  \else\ifdim\dimen255<3pt  \cellcolor{heatred!20!white}#1%
  \else\ifdim\dimen255<6pt  \cellcolor{heatred!30!white}#1%
  \else\ifdim\dimen255<10pt \cellcolor{heatred!45!white}#1%
  \else\ifdim\dimen255<15pt \cellcolor{heatred!60!white}#1%
  \else\ifdim\dimen255<22pt \cellcolor{heatred!75!white}#1%
  \else\ifdim\dimen255<30pt \cellcolor{heatred!85!white}#1%
  \else                     \cellcolor{heatred!90!white}#1%
  \fi\fi\fi\fi\fi\fi\fi
}

\newtcolorbox{promptbox}[1]{
    colback=gray!5,    
    colframe=gray!75!black, 
    fonttitle=\bfseries,  
    title=#1,         
    arc=0mm,  
    left=5pt, right=5pt, top=5pt, bottom=5pt,
    enhanced,
    breakable 
}

\usepackage{lineno}

\definecolor{darkblue}{rgb}{0, 0, 0.5}
\hypersetup{colorlinks=true, citecolor=darkblue, linkcolor=darkblue, urlcolor=darkblue}
\definecolor{MyOrange}{RGB}{252,141,98}

\usepackage{colortbl} 

\definecolor{dimmgreen}{rgb}{0, 0.7, 0} 
\newcommand{\heatcell}[1]{%
    \if\relax\detokenize{#1}\relax 
        #1
    \else
        \ifdim #1pt<10pt\cellcolor{white}\else
        \ifdim #1pt<20pt\cellcolor{dimmgreen!5!white}\else
        \ifdim #1pt<30pt\cellcolor{dimmgreen!10!white}\else
        \ifdim #1pt<40pt\cellcolor{dimmgreen!15!white}\else
        \ifdim #1pt<50pt\cellcolor{dimmgreen!25!white}\else
        \ifdim #1pt<60pt\cellcolor{dimmgreen!35!white}\else
        \ifdim #1pt<70pt\cellcolor{dimmgreen!45!white}\else
        \ifdim #1pt<75pt\cellcolor{dimmgreen!55!white}\else
        \ifdim #1pt<80pt\cellcolor{dimmgreen!65!white}\else
        \ifdim #1pt<82pt\cellcolor{dimmgreen!67!white}\else
        \ifdim #1pt<84pt\cellcolor{dimmgreen!71!white}\else
        \ifdim #1pt<86pt\cellcolor{dimmgreen!75!white}\else
        \ifdim #1pt<88pt\cellcolor{dimmgreen!79!white}\else
        \ifdim #1pt<90pt\cellcolor{dimmgreen!83!white}\else
        \ifdim #1pt<92pt\cellcolor{dimmgreen!87!white}\else
        \ifdim #1pt<94pt\cellcolor{dimmgreen!91!white}\else
        \ifdim #1pt<96pt\cellcolor{dimmgreen!95!white}\else
        \ifdim #1pt<96pt\cellcolor{dimmgreen!99!white}\else
        \cellcolor{dimmgreen!100!white}\fi\fi\fi\fi\fi\fi\fi\fi\fi\fi\fi\fi\fi\fi\fi\fi\fi\fi#1%
    \fi
}
\usepackage{subcaption}

\title{MGSM-Pro: A Simple Strategy for Robust Multilingual Mathematical Reasoning Evaluation}

\author{
  \textbf{Tianyi Xu\textsuperscript{1,2}},
  \textbf{Kosei Uemura\textsuperscript{2,3}},
  \textbf{Alfred Malengo Kondoro\textsuperscript{4*}}, \textbf{Tadesse Destaw Belay\textsuperscript{5*}},
\\
  \textbf{Catherine Nana Nyaah Essuman\textsuperscript{6*}}, \textbf{Ifeoma Okoh\textsuperscript{*}}, \textbf{Ganiyat Afolabi\textsuperscript{7*}}, 
\\
\textbf{Ayodele Awokoya\textsuperscript{7,8*}},
  \textbf{David Ifeoluwa Adelani\textsuperscript{1,2,9*}}
\\
  \textsuperscript{1}McGill University,
  \textsuperscript{2}Mila-Quebec AI Institute, \textsuperscript{3}University of Toronto, \textsuperscript{*}Masakhane,
\\
  \textsuperscript{4}Hanyang University, Rep. of Korea,  
  \textsuperscript{5}Instituto Politécnico Nacional, Mexico,  \textsuperscript{6}Umbaji,\\
  \textsuperscript{7}University of Ibadan, Nigeria,  \textsuperscript{8}McPherson University, Nigeria,  
  \textsuperscript{9}Canada CIFAR AI Chair
\\
}

\begin{document}
\maketitle
\begin{abstract}

Large language models have made substantial progress in mathematical reasoning. However, benchmark development for multilingual evaluation has lagged behind English in both difficulty and recency. Recently, GSM-Symbolic~\citep{mirzadeh2025gsmsymbolic} showed a strong evidence of high variance when models are evaluated on different instantiations of the same question; however, the evaluation was conducted only in English. In this paper, we introduce MGSM-Pro, an extension of MGSM dataset with GSM-Symbolic approach. Our dataset provides five instantiations per MGSM question by varying names,  digits and irrelevant context. Evaluations across nine languages reveal that many low-resource languages suffer large performance drops when tested on digit instantiations different from those in the original test set. We further find that models robustness in HRL setting do not necessarily translate to LRL. Moreover, proprietary models, such as Gemini 2.5 Flash and GPT-4.1 are less robust to digit, whereas Gemini 3.0 Pro is more robust. Among open models, GPT-OSS 120B and DeepSeek v3 show stronger robustness. Based on these findings, we recommend evaluating each problem using at least five digit-varying instantiations to obtain a more robust and realistic assessment of math reasoning.

\end{abstract}

\section{Introduction}

Large language models (LLMs) have drastically improved in capability in recent years, particularly on challenging knowledge-intensive and reasoning tasks, with open models closing the gap as evidenced by public benchmarks~\citep{liu2024deepseek,yang2025qwen3,team2025gemma}. However, progress in developing benchmarks for multilingual settings, particularly for mathematical reasoning, has lagged behind English in both difficulty and recency,~\footnote{E.g. AIME \url{https://artofproblemsolving.com/wiki/index.php/AIME_Problems_and_Solutions}} making existing multilingual benchmarks easily saturated and potentially prone to memorization or being over-optimized~\citep{shilanguage,chen-etal-2024-breaking}.

\begin{figure*}[t]
    \centering
    \includegraphics[width=0.99\linewidth]{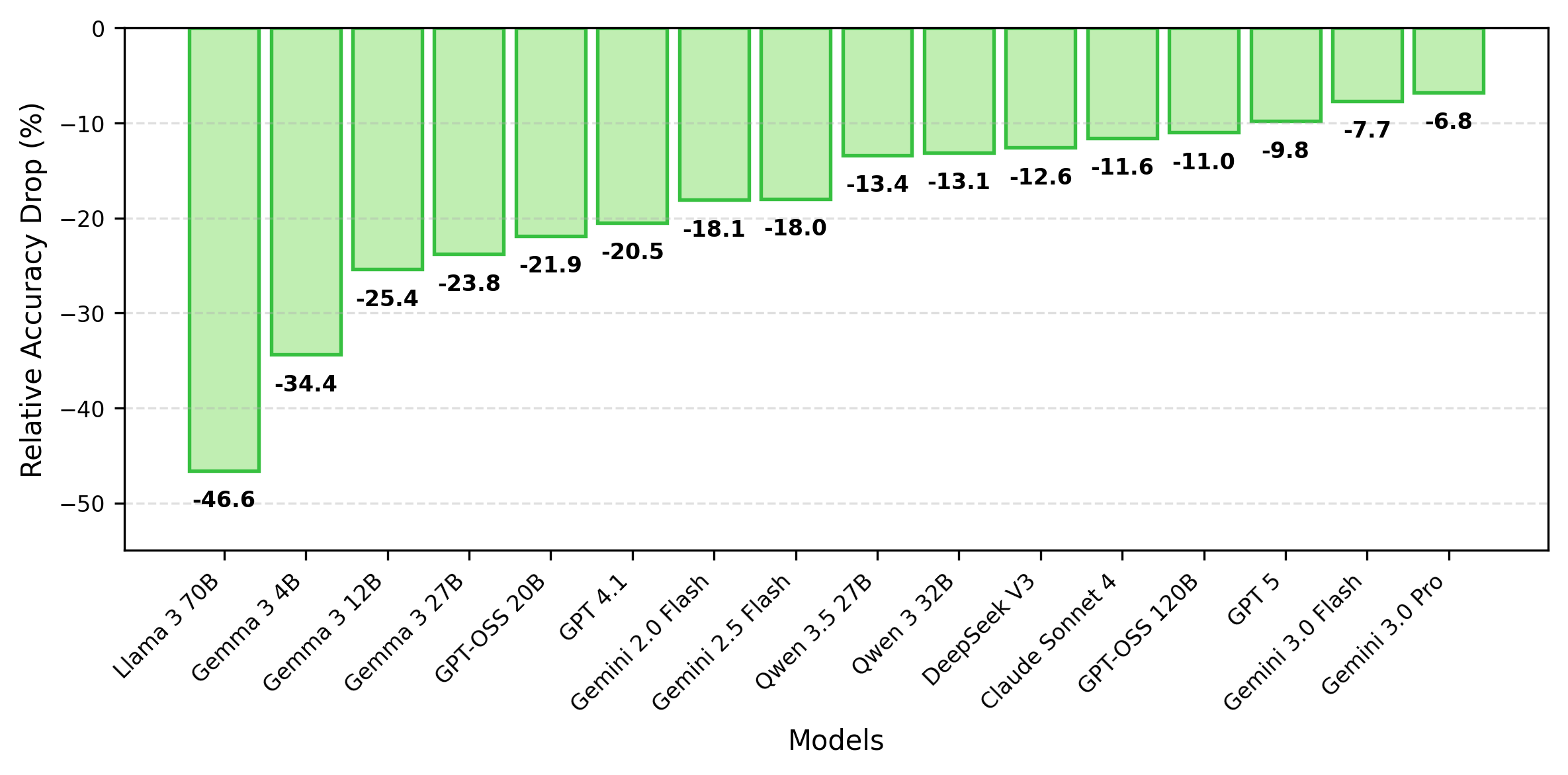}
    \vspace{-2mm}
    \caption{Relative decrease in accuracy from the original dataset to five instances of changing both names and numbers and adding irrelevant context, averaged over all nine languages.}
    \label{fig:relative_dataset_accuracy_drop}
\end{figure*}

One way to address this issue is to create new benchmarks that are more recent, such as MMath~\citep{luo-etal-2025-mmath} and PolyMath~\citep{wang2025polymath} often translated from existing English benchmarks, but without modification of the numbers and context. However, it remains unclear whether LLMs evaluated on these benchmarks generalize to other similar problems.
Prior evidence in English shows that LLMs exhibit high variance when presented with different instantiations of the same question (known as GSM-Symbolic)~\citep{mirzadeh2025gsmsymbolic}. We carefully extend this finding to the multilingual setting.

In this paper, we introduce \textbf{\texttt{MGSM-Pro}}, a multilingual extension of \texttt{GSM-Symbolic} based on the MGSM and AfriMGSM datasets~\citep{shi-etal-2022,adelani-etal-2025}  in two steps: (1) \textit{template construction} in English that allows easy replacement of names and digits (2) \textit{dataset construction} that translates the template to multiple languages (with an LLM), followed by human verification---this helps to generate different instantiations of same question (e.g. 5 instances). 

Our results reveal a more precarious setting than GSM-Symbolic: low-resource languages (LRLs) experience a sharp performance drop when accuracy is averaged over five instances instead of a single example, unlike high-resource languages (HRLs). As shown in \autoref{fig:relative_dataset_accuracy_drop}, Gemini 3.0 Pro and Gemini 3.0 Flash are more robust to this degradation, whereas smaller-sized open models like Gemma 3 4B and older LLMs (regardless of size) such as Llama 3 70B struggle to maintain accuracy relative to the original dataset. When languages are grouped by resource level (i.e. HRL vs. LRLs), LRLs suffer the most in terms of a huge drop in performance and, in some cases, lose more than a $-20.0$ drop in performance. Finally, we show that leaderboard rankings can undergo substantial changes when results are averaged over at least five instances, with Gemini 2.5 Flash, for example, falling from 3rd to 7th place.

Based on our findings on nine typologically diverse languages, \textbf{we recommend that math reasoning evaluation should be performed on a minimum of five instances} of the same problem by modifying digits.\footnote{i.e. evaluating on 1240 instances of MGSM rather than the 248 questions for a more robust evaluation}  We are \textbf{\textit{releasing the new dataset} (\texttt{MGSM-Pro})} with more instances to encourage a more robust evaluation. Similar to how we expect a good student that understands a sample problem to be able to solve various instances with modified digits. We expect both open LLMs and proprietary LLMs to be robust to these small changes.  The dataset will be released on HuggingFace on paper acceptance.
\begin{figure*}[t]
    \centering
    \includegraphics[width=0.8\linewidth]{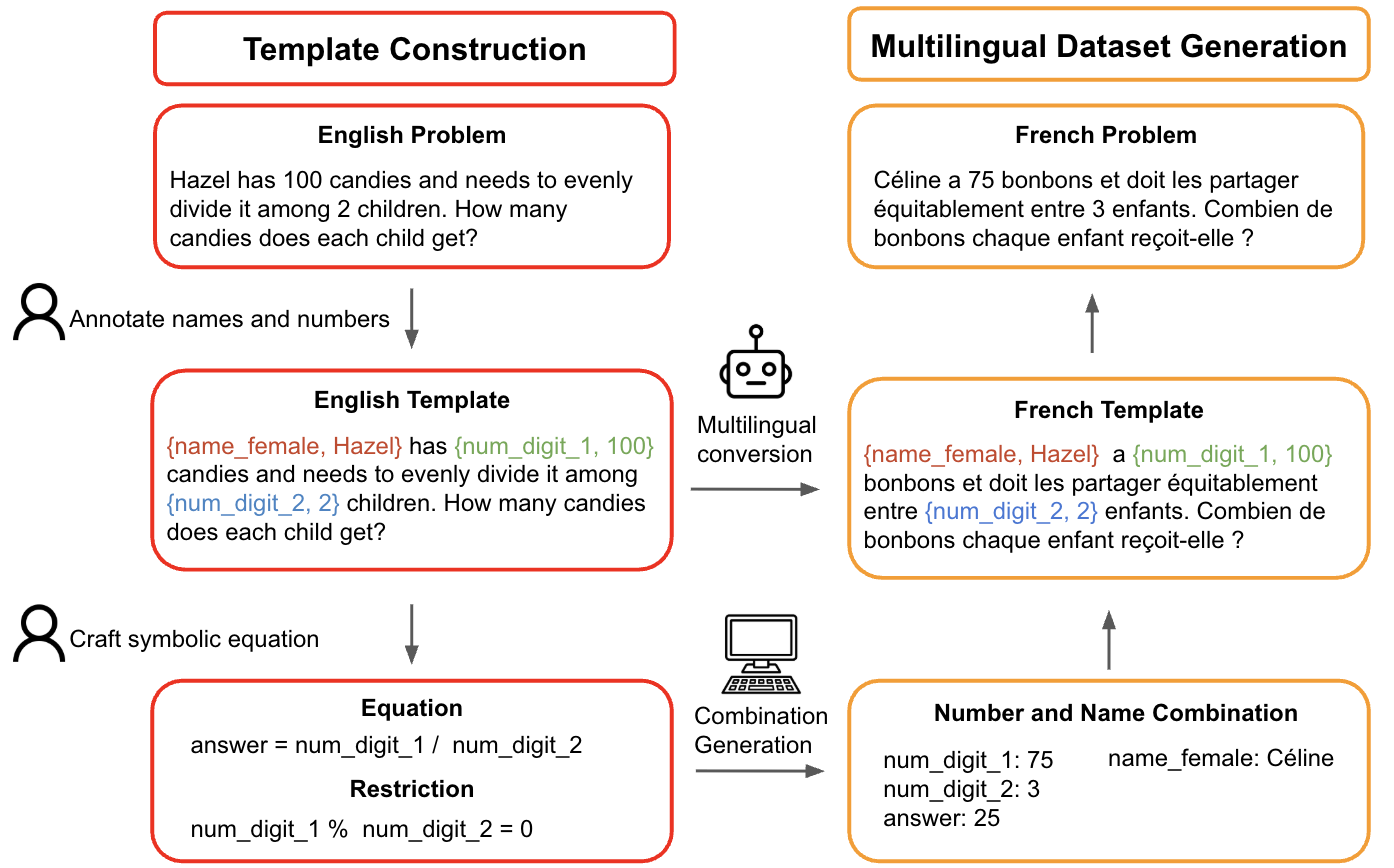}
    \vspace{-2mm}
    \caption{MGSM-Pro creation diagram, illustrating both template creation and multilingual data construction steps on a sample language.}
    \label{fig:workflow}
\end{figure*}
\section{Related Work}

\textbf{Math Reasoning Benchmarks}
With the increase of interest in evaluating a model's logical reasoning capabilities, multiple English math benchmarks have been introduced \citep{cobbe-etal-2021, hendrycks-etal-2021, mishra-etal-2022, patel-etal-2021, miao-etal-2020}. Extending the investigation into the multilingual setting, \citep{shi-etal-2022, adelani-etal-2025} notices weaker model performances under low-resource language settings. However, it is unclear if success on these benchmarks translates to effectiveness on related problems or memorization of the test set.

\textbf{Robustness in Reasoning}
True logical reasoning requires robustness to minor variations and noise. Several English datasets highlight significant accuracy drops in such scenarios \citep{shi-etal-2023, abedin-etal-2025, mirzadeh2025gsmsymbolic,adelani-etal-2025}. However, their investigations remain limited to English. Our work introduces \texttt{MGSM-Pro}, a new dataset that expands these investigations to a multilingual setting.

\textbf{Re-purposing existing benchmarks} Scaling labeled datasets across many languages remains challenging due to annotation costs and the difficulty of constructing sufficiently challenging benchmarks. Recent work has explored re-purposing existing datasets to increase both their complexity and coverage. For instance, SIB-200~\citep{adelani-etal-2024-sib} and Belebele~\citep{bandarkar-etal-2024-belebele} extend the FLORES-200 benchmark by introducing additional labels or multiple-choice formulations, enabling evaluation across a broader set of languages. Similarly, MMLU-Pro~\cite{wang2024mmlupro} increases task difficulty of MMLU~\citep{hendrycks2021measuring} by expanding the number of answer choices from four to eight, while MMLU-ProX~\citep{xuan-etal-2025-mmlu} further extends this framework to additional languages. GlobalMMLU augments MMLU with annotations that distinguish between questions requiring Western cultural knowledge and those that do not. Collectively, these efforts enable more rigorous and scalable evaluation of large language models across diverse languages and tasks. Building on this line of work, we introduce \texttt{MGSM-Pro}, which expands the original MGSM dataset fivefold (248 to 1,240 questions) by systematically generating new instances through controlled digit substitutions. This approach enables a more robust and fine-grained evaluation of multilingual mathematical reasoning.

\section{MGSM-Pro: Creation Process}
We introduce, \textbf{\texttt{MGSM-Pro}}---a multilingual extension of \texttt{GSM-Symbolic} based on the MGSM and AfriMGSM datasets~\citep{shi-etal-2022,adelani-etal-2025} to nine languages with various resource levels as defined by \citet{joshi-etal-2020-state}. This includes high-resource languages or HRLs (\textbf{English}, \textbf{Chinese}, \textbf{French}, and \textbf{Japanese}; Class 5) and low-resource languages or LRLs (\textbf{Swahili}, \textbf{Amharic}, \textbf{Igbo}, \textbf{Yoruba}, and \textbf{Twi}; Classes 1–2). We also cover six dataset variants per language. These variations are organized into two series: Symbolic (\sym) and Irrelevant Context (\ic). Each series consists of three distinct variations.

\textbf{The Symbolic Series (\sym)} involves systematic modifications to a problem's surface features without altering its logical structure. This series includes three variants: 
\begin{itemize}
    \item \textbf{\symn}, which replaces \textit{names} with culturally relevant ones; 
    \item \textbf{\symhash}, which changes \textit{numerical} data; and 
    \item \textbf{\symboth}, which varies \textit{both names and numbers} simultaneously.
\end{itemize}

\textbf{The Irrelevant Context Series (\ic)} mirrors the modifications in the \textbf{\sym} series but introduces a distinct layer of difficulty as it inserts an \textit{irrelevant sentence} to the problem. The resulting variants are denoted as \textbf{\icn}, \textbf{\ichash}, and \textbf{\icboth}.

In this section, we introduce the methodology for constructing MGSM-Pro in two steps: template construction (§\ref{subsec:template_construction}) and dataset construction (§\ref{subsec:dataset_construction}). \autoref{fig:workflow} shows an example of the data generation workflow in which names and digits are first identified and replaced with multiple instances.  

\subsection{Template Construction} 
\label{subsec:template_construction}
The foundation of our dataset lies in the creation of adaptable templates. We adopt the GSM-Symbolic framework to generate symbolic templates for 248 out of 250 English MGSM questions.\footnote{The remaining two were excluded because the questions do not have digits, so it is difficult to use a template approach that focuses on digit replacement.} To simplify cross-lingual transfer, we restrict parameterization strictly to names and numbers (i.e. \sym). Each template includes a symbolic equation alongside variable constraints to ensure that the generated combinations yield correct and logical answers. Once the English template is crafted, we employ Gemini 2.0 Flash to generate multilingual templates. These translations then undergo a rigorous verification process: they are first reviewed by native speakers, followed by automated alignment checks against the English source. Any template failing these checks is subject to a second round of human correction. Finally, to enable a controlled increase in difficulty, we build upon GSM-IC's~\citep{mirzadeh2025gsmsymbolic} methodology to create irrelevant context templates for every English question. The curation of the IC sentence template follows the methodology of \cite{shi-etal-2023}, where we ensure that irrelevant sentences have: 1) some related connection with the problem and 2) use names found in the question. We applied similar rigorous checks as with the \sym templates. More details on the template construction process are in Appendix \ref{app:annotation_protocol}.

\subsection{Dataset Construction}
\label{subsec:dataset_construction}
To efficiently generate a large quantity of problem instances that share the same underlying logical structure, we leverage the symbolic equations and restrictions defined during the template phase. This methodology enables the systematic sampling of new numerical values that are guaranteed to be mathematically valid and distinct from those present in the original training data.

A limitation in previous datasets, such as MGSM and AfriMGSM, was the reliance on direct translations, where names were frequently phonetic transliterations of English origin. This approach compromised the problems' local fit and cultural meaning. To ensure deep cultural relevance across all languages in MGSM-Pro, we tasked native annotators with curating a comprehensive repository of entities specific to their locale. This includes categories such as cities, personal names, and common pet names, guaranteeing that the generated problems resonate well with native speakers and accurately represent the target language's culture. For example an annotator suggested using 'Zainabu' as a female name in the Swahili dataset as opposed to 'Carla' which was found in the original Swahili MGSM dataset.

\begin{table*}[t]
  \centering
  \footnotesize
  \setlength{\tabcolsep}{3pt}
  \renewcommand{\arraystretch}{1.15}
  \resizebox{\textwidth}{!}{%
  \begin{tabular}{l cccc !{\color{black!30}\vrule} cccc !{\color{black!30}\vrule} cccc !{\color{black!30}\vrule} cccc !{\color{black!30}\vrule} cccc !{\color{black!30}\vrule} cc}
    \toprule
    & \multicolumn{4}{c}{\textbf{Gemini 2.5 Flash}}
    & \multicolumn{4}{c}{\textbf{Gemini 3.0 Pro}}
    & \multicolumn{4}{c}{\textbf{Claude 4 Sonnet}}
    & \multicolumn{4}{c}{\textbf{GPT-4.1}}
    & \multicolumn{4}{c}{\textbf{GPT-5}}
    & $\Delta$ Ave. & $\Delta$ Med. \\
    \cmidrule(lr){2-5} \cmidrule(lr){6-9} \cmidrule(lr){10-13} \cmidrule(lr){14-17} \cmidrule(lr){18-21} \cmidrule(lr){22-23}
    \textbf{Language}
    & $D_O$ & \icn & \symhash & \ichash
    & $D_O$ & \icn & \symhash & \ichash
    & $D_O$ & \icn & \symhash & \ichash
    & $D_O$ & \icn & \symhash & \ichash
    & $D_O$ & \icn & \symhash & \ichash
    & -- & -- \\
    \midrule
    English  & \dr{96.8} & \hc{94.6} & \hc{83.1} & \hc{81.0} & \dr{98.0} & \hc{96.2} & \hc{94.4} & \hc{93.2} & \dr{98.0} & \hc{96.0} & \hc{91.5} & \hc{90.4} & \dr{96.4} & \hc{91.7} & \hc{81.5} & \hc{79.6} & \dr{96.8} & \hc{93.3} & \hc{92.8} & \hc{88.0} & $-10.7$ & $-8.8$  \\
    Chinese  & \dr{89.9} & \hc{89.0} & \hc{78.0} & \hc{79.0} & \dr{93.5} & \hc{93.1} & \hc{93.2} & \hc{93.5} & \dr{93.1} & \hc{91.9} & \hc{88.7} & \hc{88.5} & \dr{89.9} & \hc{90.6} & \hc{78.6} & \hc{76.8} & \dr{92.3} & \hc{91.9} & \hc{89.3} & \hc{88.4} & $-6.5$  & $-4.7$  \\
    French   & \dr{91.5} & \hc{87.1} & \hc{77.5} & \hc{73.8} & \dr{90.7} & \hc{89.7} & \hc{88.6} & \hc{87.2} & \dr{91.5} & \hc{90.4} & \hc{86.4} & \hc{85.3} & \dr{88.7} & \hc{85.5} & \hc{75.9} & \hc{74.5} & \dr{89.9} & \hc{88.4} & \hc{85.4} & \hc{84.2} & $-9.5$  & $-6.2$  \\
    Japanese & \dr{86.7} & \hc{83.9} & \hc{74.9} & \hc{73.1} & \dr{90.7} & \hc{89.4} & \hc{87.6} & \hc{88.0} & \dr{89.9} & \hc{84.8} & \hc{83.1} & \hc{81.5} & \dr{87.1} & \hc{83.6} & \hc{74.8} & \hc{74.1} & \dr{90.7} & \hc{84.1} & \hc{83.6} & \hc{82.6} & $-9.2$  & $-8.5$  \\
    Swahili  & \dr{91.5} & \hc{89.9} & \hc{80.3} & \hc{78.7} & \dr{97.6} & \hc{93.9} & \hc{93.0} & \hc{92.4} & \dr{91.9} & \hc{90.9} & \hc{85.1} & \hc{84.4} & \dr{91.5} & \hc{89.0} & \hc{79.8} & \hc{77.5} & \dr{90.7} & \hc{92.4} & \hc{87.7} & \hc{89.3} & $-8.2$  & $-7.5$  \\
    Amharic  & \dr{81.9} & \hc{81.9} & \hc{71.0} & \hc{70.2} & \dr{87.5} & \hc{85.4} & \hc{83.2} & \hc{83.0} & \dr{82.3} & \hc{81.5} & \hc{74.5} & \hc{73.1} & \dr{68.1} & \hc{68.2} & \hc{53.0} & \hc{51.2} & \dr{71.8} & \hc{74.0} & \hc{67.6} & \hc{70.1} & $-8.8$  & $-9.2$  \\
    Igbo     & \dr{81.5} & \hc{79.7} & \hc{68.6} & \hc{66.6} & \dr{89.9} & \hc{87.5} & \hc{82.1} & \hc{81.2} & \dr{78.2} & \hc{76.7} & \hc{68.1} & \hc{66.0} & \dr{79.0} & \hc{71.3} & \hc{58.5} & \hc{54.0} & \dr{79.8} & \hc{74.5} & \hc{70.2} & \hc{66.7} & $-14.8$ & $-13.1$ \\
    Yoruba   & \dr{83.9} & \hc{79.4} & \hc{69.7} & \hc{68.1} & \dr{86.3} & \hc{85.7} & \hc{83.1} & \hc{80.9} & \dr{77.4} & \hc{73.7} & \hc{67.9} & \hc{66.0} & \dr{72.6} & \hc{69.6} & \hc{59.8} & \hc{55.6} & \dr{74.6} & \hc{73.5} & \hc{67.7} & \hc{68.9} & $-11.2$ & $-11.4$ \\
    Twi      & \dr{65.3} & \hc{61.9} & \hc{51.5} & \hc{49.6} & \dr{77.8} & \hc{74.9} & \hc{70.6} & \hc{69.1} & \dr{52.8} & \hc{46.0} & \hc{43.2} & \hc{36.2} & \dr{41.9} & \hc{36.9} & \hc{31.5} & \hc{28.4} & \dr{46.8} & \hc{43.1} & \hc{38.5} & \hc{37.9} & $-12.7$ & $-13.5$ \\
    \midrule
    Average  & \dr{85.4} & \hc{83.0} & \hc{72.7} & \hc{71.1} & \dr{90.2} & \hc{88.4} & \hc{86.2} & \hc{85.4} & \dr{83.9} & \hc{81.3} & \hc{76.5} & \hc{74.6} & \dr{79.5} & \hc{76.3} & \hc{65.9} & \hc{63.5} & \dr{81.5} & \hc{79.5} & \hc{75.9} & \hc{75.0} & & \\
    \bottomrule
  \end{tabular}%
  }
  \caption{Different closed models' accuracy across dataset variations (\symhash, \icn, \ichash) and original ($D_O$). Cells are shaded by how far each variant falls below its row's $D_O$ within the same model group; values at or above $D_O$ are left white. We report the Average (Ave.) and Median (Med.) of $\Delta(\ichash - D_O)$ taken across the five models.}
  \label{tab:main_result_closed_models}
\end{table*}
\begin{table*}[t]
  \centering
  \footnotesize
  \setlength{\tabcolsep}{3pt}
  \renewcommand{\arraystretch}{1.15}
  \resizebox{\textwidth}{!}{%
  \begin{tabular}{l cccc !{\color{black!30}\vrule} cccc !{\color{black!30}\vrule} cccc !{\color{black!30}\vrule} cccc !{\color{black!30}\vrule} cccc !{\color{black!30}\vrule} cc}
    \toprule
    & \multicolumn{4}{c}{\textbf{Gemma 3 27B}}
    & \multicolumn{4}{c}{\textbf{Qwen 3 32B}}
    & \multicolumn{4}{c}{\textbf{Qwen 3.5 27B}}
    & \multicolumn{4}{c}{\textbf{DeepSeek V3}}
    & \multicolumn{4}{c}{\textbf{GPT-OSS 120B}}
    & $\Delta$ Ave. & $\Delta$ Med. \\
    \cmidrule(lr){2-5} \cmidrule(lr){6-9} \cmidrule(lr){10-13} \cmidrule(lr){14-17} \cmidrule(lr){18-21} \cmidrule(lr){22-23}
    \textbf{Language}
    & $D_O$ & \icn & \symhash & \ichash
    & $D_O$ & \icn & \symhash & \ichash
    & $D_O$ & \icn & \symhash & \ichash
    & $D_O$ & \icn & \symhash & \ichash
    & $D_O$ & \icn & \symhash & \ichash
    & -- & -- \\
    \midrule
    English  & \dr{95.6} & \hc{93.3} & \hc{81.6} & \hc{77.6} & \dr{89.1} & \hc{87.3} & \hc{87.7} & \hc{86.5} & \dr{98.4} & \hc{94.2} & \hc{93.8} & \hc{90.7} & \dr{98.4} & \hc{94.3} & \hc{92.7} & \hc{89.7} & \dr{96.4} & \hc{93.9} & \hc{93.5} & \hc{92.1} & $-8.3$  & $-7.7$  \\
    Chinese  & \dr{87.9} & \hc{85.5} & \hc{75.2} & \hc{72.3} & \dr{89.9} & \hc{88.8} & \hc{87.8} & \hc{88.1} & \dr{89.9} & \hc{90.1} & \hc{87.3} & \hc{83.9} & \dr{92.3} & \hc{91.4} & \hc{89.9} & \hc{88.1} & \dr{91.5} & \hc{90.6} & \hc{90.2} & \hc{89.7} & $-5.9$  & $-4.2$  \\
    French   & \dr{89.1} & \hc{84.5} & \hc{73.4} & \hc{69.9} & \dr{89.5} & \hc{87.6} & \hc{86.3} & \hc{84.0} & \dr{91.1} & \hc{87.2} & \hc{87.2} & \hc{84.1} & \dr{90.7} & \hc{88.7} & \hc{87.8} & \hc{85.7} & \dr{90.7} & \hc{88.7} & \hc{86.5} & \hc{86.5} & $-8.2$  & $-5.5$  \\
    Japanese & \dr{85.1} & \hc{79.4} & \hc{71.1} & \hc{64.6} & \dr{88.7} & \hc{85.2} & \hc{84.8} & \hc{83.3} & \dr{87.5} & \hc{80.1} & \hc{76.0} & \hc{72.1} & \dr{88.7} & \hc{83.0} & \hc{79.8} & \hc{79.8} & \dr{89.1} & \hc{85.2} & \hc{85.2} & \hc{83.9} & $-11.1$ & $-8.9$  \\
    Swahili  & \dr{89.1} & \hc{85.4} & \hc{72.2} & \hc{71.2} & \dr{78.2} & \hc{71.0} & \hc{73.6} & \hc{67.3} & \dr{91.9} & \hc{88.5} & \hc{88.1} & \hc{84.9} & \dr{90.7} & \hc{89.6} & \hc{86.3} & \hc{84.0} & \dr{84.7} & \hc{82.2} & \hc{81.5} & \hc{79.6} & $-9.5$  & $-7.0$  \\
    Amharic  & \dr{71.4} & \hc{70.2} & \hc{57.4} & \hc{55.1} & \dr{42.7} & \hc{39.4} & \hc{37.7} & \hc{33.6} & \dr{77.0} & \hc{76.0} & \hc{68.1} & \hc{67.4} & \dr{76.2} & \hc{73.5} & \hc{71.9} & \hc{66.0} & \dr{60.9} & \hc{57.7} & \hc{58.5} & \hc{53.7} & $-10.5$ & $-9.6$  \\
    Igbo     & \dr{64.1} & \hc{54.0} & \hc{50.0} & \hc{41.3} & \dr{19.8} & \hc{14.9} & \hc{14.7} & \hc{10.0} & \dr{73.8} & \hc{70.8} & \hc{69.5} & \hc{62.9} & \dr{73.8} & \hc{67.8} & \hc{64.6} & \hc{58.0} & \dr{77.4} & \hc{73.3} & \hc{67.7} & \hc{64.7} & $-14.4$ & $-12.7$ \\
    Yoruba   & \dr{44.4} & \hc{40.5} & \hc{38.2} & \hc{31.4} & \dr{26.6} & \hc{17.3} & \hc{23.1} & \hc{14.0} & \dr{74.6} & \hc{70.0} & \hc{69.3} & \hc{63.2} & \dr{66.1} & \hc{58.0} & \hc{57.1} & \hc{52.7} & \dr{76.2} & \hc{66.9} & \hc{66.1} & \hc{61.3} & $-13.0$ & $-13.0$ \\
    Twi      & \dr{18.1} & \hc{10.4} & \hc{13.4} & \hc{9.3}  & \dr{5.6}  & \hc{3.5}  & \hc{4.6}  & \hc{2.2}  & \dr{37.5} & \hc{27.4} & \hc{31.3} & \hc{25.4} & \dr{48.0} & \hc{36.9} & \hc{38.0} & \hc{28.1} & \dr{44.8} & \hc{37.7} & \hc{39.0} & \hc{31.5} & $-11.5$ & $-12.1$ \\
    \midrule
    Average  & \dr{71.6} & \hc{67.0} & \hc{59.2} & \hc{54.7} & \dr{58.9} & \hc{55.0} & \hc{55.6} & \hc{52.1} & \dr{80.2} & \hc{76.0} & \hc{74.5} & \hc{70.5} & \dr{80.6} & \hc{75.9} & \hc{74.2} & \hc{70.3} & \dr{79.1} & \hc{75.1} & \hc{74.2} & \hc{71.4} & & \\
    \bottomrule
  \end{tabular}%
  }
  \caption{Different open models' accuracy across dataset variations (\symhash, \icn, \ichash) and original ($D_O$). Cells are shaded by how far each variant falls below its row's $D_O$ within the same model group; values at or above $D_O$ are left white. We report the Average (Ave.) and Median (Med.) of $\Delta(\ichash - D_O)$ taken across the five models.}
  \label{tab:main_result_open_models}
\end{table*}

\begin{figure*}[t]
    \centering
    \includegraphics[width=0.99\linewidth]{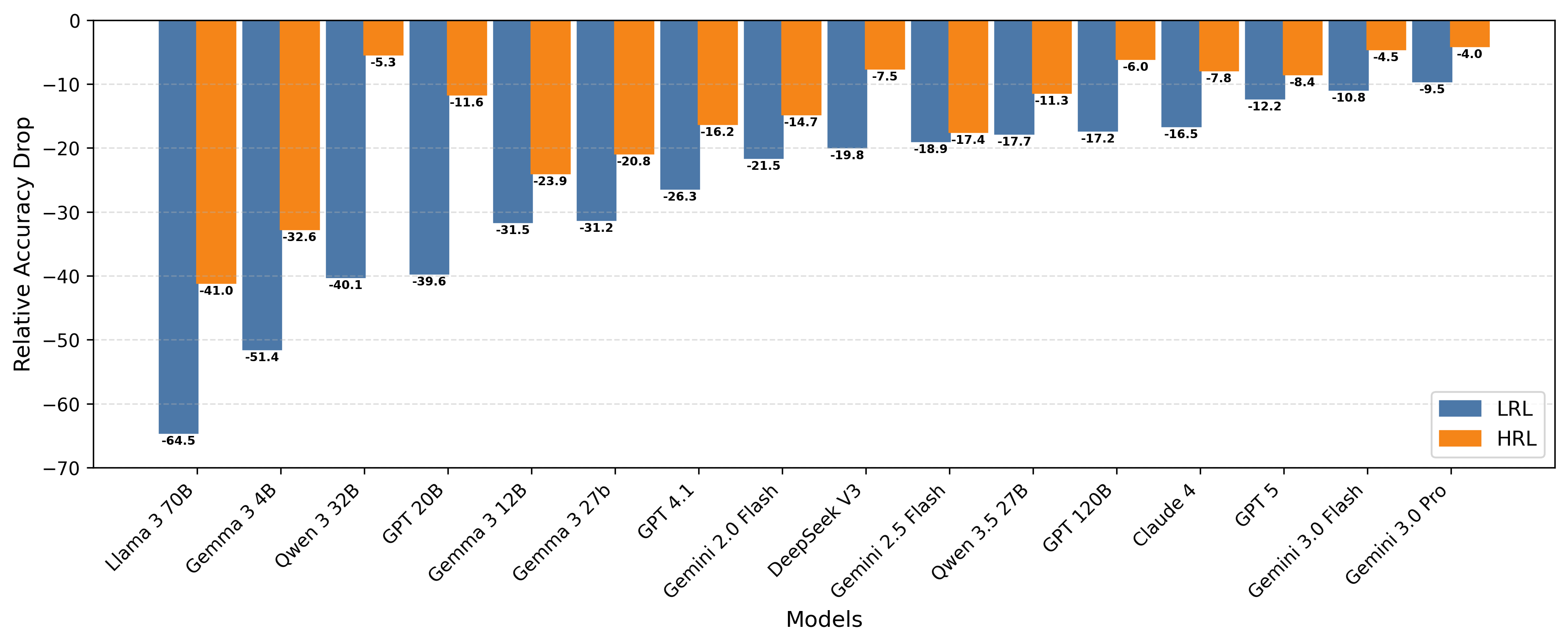}
    \vspace{-2mm}
    \caption{Comparison of relative accuracy decrease from HRL and LRL $D_o$, averaged across six variants of MGSM-Pro. Models are sorted in descending order of average accuracy drop.}
    \label{fig:lrl_vs_hrl_relative_decrease}
\end{figure*}

\begin{figure*}[t]
    \centering

    \begin{subfigure}[b]{0.4\linewidth}
        \centering
        \includegraphics[width=\linewidth]{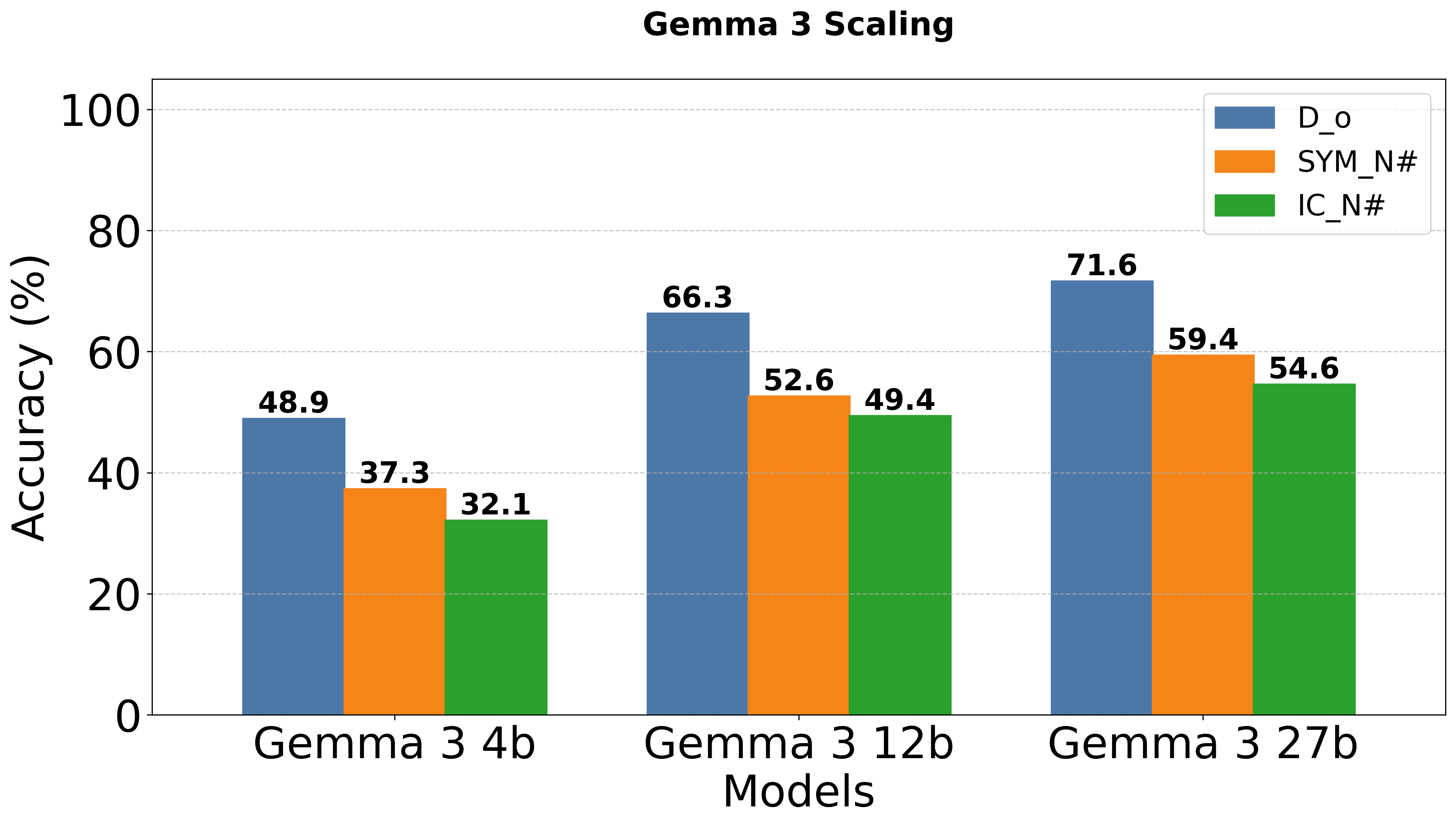}
        \caption{Gemma 3 Family}
        \label{fig:gemma_comparison}
    \end{subfigure}
    \hspace{1cm}
    \begin{subfigure}[b]{0.4\linewidth}
        \centering
        \includegraphics[width=\linewidth]{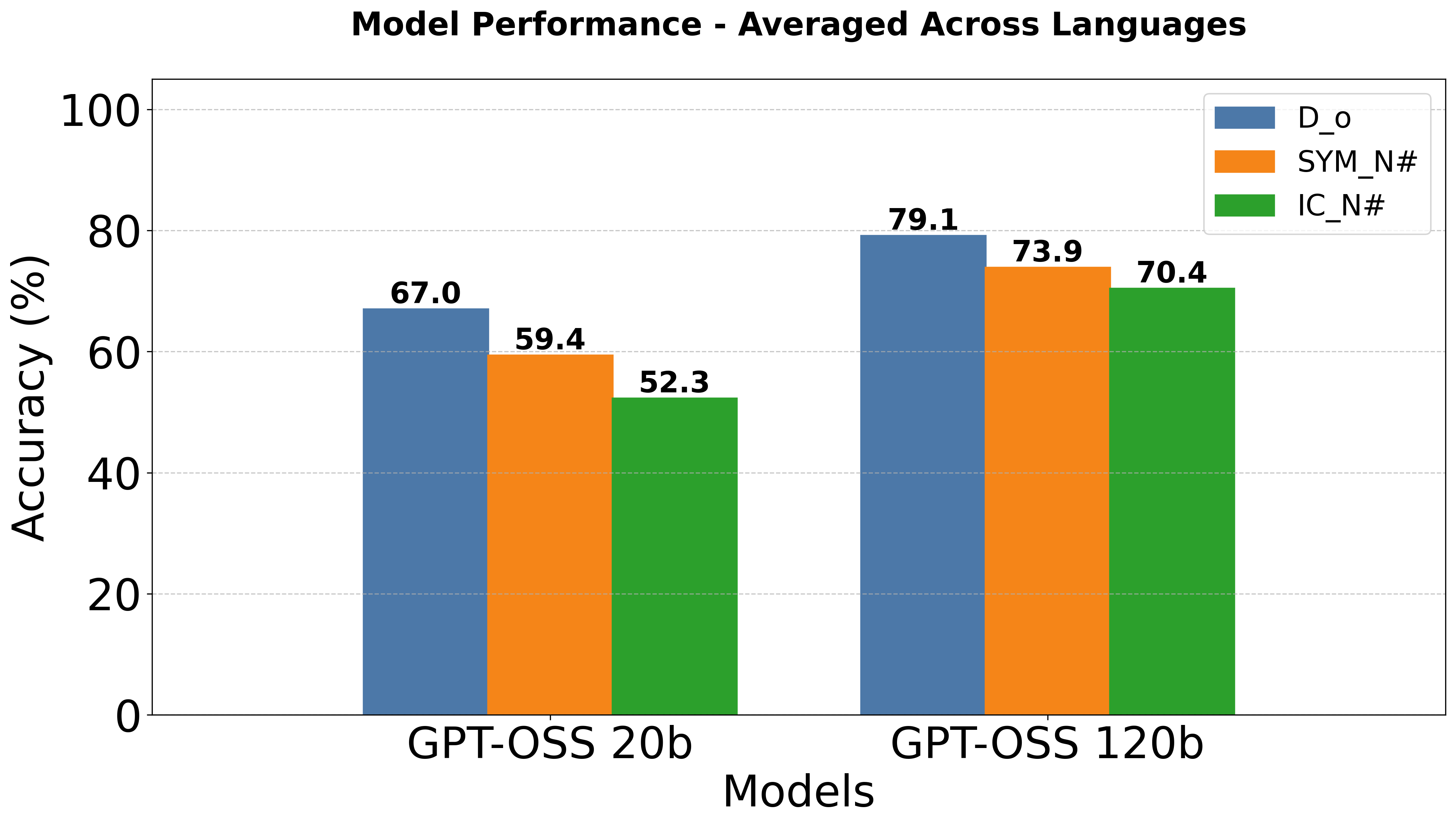}
        \caption{GPT-OSS Family}
        \label{fig:gpt_oss_comparison}
    \end{subfigure}

    \vspace{-2mm}

    \caption{\textbf{Relative Accuracy Drop Across Model Families.} The figures illustrate the relative decline in accuracy for (a) the Gemma-3 family and (b) the GPT-OSS family. The drop is measured from the original dataset to two configurations: IC\_N\# and SYM\_N\#, averaged over nine languages.}
    \label{fig:combined_model_scaling_comparison}
\end{figure*}

\begin{table*}[t]
  \centering
  \scriptsize
  \setlength{\tabcolsep}{2.5pt}
  \renewcommand{\arraystretch}{1.15}
  \begin{tabular}{l cc !{\color{black!30}\vrule} cccc !{\color{black!30}\vrule} cccc !{\color{black!30}\vrule} cccc !{\color{black!30}\vrule} ccc !{\color{black!30}\vrule} ccc}
    \toprule
    & \multicolumn{14}{c}{\textbf{All Language}}
    & \multicolumn{3}{c}{\textbf{High-Resource}}
    & \multicolumn{3}{c}{\textbf{Low-Resource}} \\
    \cmidrule(lr){2-15} \cmidrule(lr){16-18} \cmidrule(lr){19-21}
    \textbf{Model}
    & \multicolumn{2}{c}{$D_O$}
    & \multicolumn{4}{c}{Avg-3}
    & \multicolumn{4}{c}{Avg-5}
    & \multicolumn{4}{c}{Avg-10}
    & \multicolumn{3}{c}{Avg-5}
    & \multicolumn{3}{c}{Avg-5} \\
    \cmidrule(lr){2-3} \cmidrule(lr){4-7} \cmidrule(lr){8-11} \cmidrule(lr){12-15} \cmidrule(lr){16-18} \cmidrule(lr){19-21}
    & \faMedal & \faPercent
    & \multicolumn{1}{c}{\faMedal} & \faSort & \faPercent & $\pm$
    & \multicolumn{1}{c}{\faMedal} & \faSort & \faPercent & $\pm$
    & \multicolumn{1}{c}{\faMedal} & \faSort & \faPercent & $\pm$
    & \multicolumn{1}{c}{\faMedal} & \faPercent & $\pm$
    & \multicolumn{1}{c}{\faMedal} & \faPercent & $\pm$ \\
    \midrule
    Gemini 3.0 Pro    & 1  & 90.2 & 1  & \rkno & 84.2 & 2.7 & 1  & \rkno & 84.1 & 1.5 & 1  & \rkno & 84.2 & 0.9 & 1  & 89.6 & 1.3 & 1  & 79.7 & 1.6 \\
    Gemini 3.0 Flash  & 2  & 89.2 & 2  & \rkno & 82.6 & 2.3 & 2  & \rkno & 82.3 & 1.5 & 2  & \rkno & 82.4 & 0.8 & 2  & 89.5 & 1.5 & 2  & 76.5 & 1.5 \\
    Gemini 2.5 Flash  & 3  & 85.4 & 6  & \rkdn & 70.3 & 3.3 & 7  & \rkdn & 70.1 & 1.8 & 7  & \rkno & 69.6 & 1.1 & 12 & 75.4 & 1.6 & 3  & 65.8 & 2.0 \\
    Claude 4          & 4  & 83.9 & 3  & \rkup & 74.3 & 3.2 & 3  & \rkno & 74.2 & 1.5 & 3  & \rkno & 73.9 & 1.0 & 4  & 85.9 & 1.4 & 4  & 64.8 & 1.6 \\
    Gemini 2.0 Flash  & 5  & 83.4 & 9  & \rkdn & 68.4 & 4.4 & 9  & \rkno & 68.3 & 2.3 & 9  & \rkno & 67.8 & 1.5 & 10 & 76.2 & 1.8 & 6  & 62.0 & 2.7 \\
    GPT-5             & 6  & 81.5 & 5  & \rkup & 70.4 & 4.1 & 4  & \rkup & 73.3 & 2.0 & 4  & \rkno & 73.5 & 1.1 & 7  & 84.4 & 1.8 & 5  & 64.5 & 2.2 \\
    DeepSeek V3       & 7  & 80.6 & 4  & \rkup & 73.3 & 3.5 & 5  & \rkdn & 70.5 & 1.6 & 6  & \rkdn & 70.7 & 1.1 & 5  & 85.6 & 1.6 & 8  & 58.4 & 1.6 \\
    Qwen 3.5 27B      & 8  & 80.2 & 8  & \rkno & 69.3 & 4.1 & 8  & \rkno & 69.4 & 1.9 & 8  & \rkno & 69.4 & 1.1 & 8  & 81.5 & 2.0 & 7  & 59.8 & 1.9 \\
    GPT 4.1           & 9  & 79.5 & 10 & \rkdn & 63.4 & 3.5 & 10 & \rkno & 63.2 & 1.8 & 10 & \rkno & 63.1 & 1.1 & 11 & 75.9 & 1.6 & 10 & 53.1 & 1.9 \\
    GPT-OSS 120B      & 10 & 79.1 & 7  & \rkup & 70.2 & 4.0 & 6  & \rkup & 70.4 & 1.7 & 5  & \rkup & 70.8 & 1.1 & 3  & 86.4 & 1.5 & 9  & 57.6 & 1.9 \\
    Gemma 3 27B       & 11 & 71.6 & 11 & \rkno & 54.4 & 4.3 & 11 & \rkno & 54.6 & 2.0 & 11 & \rkno & 53.9 & 1.5 & 13 & 70.8 & 2.0 & 11 & 41.6 & 2.0 \\
    GPT-OSS 20B       & 12 & 67.0 & 12 & \rkno & 52.3 & 3.4 & 12 & \rkno & 52.3 & 1.8 & 12 & \rkno & 52.5 & 1.1 & 9  & 78.9 & 1.5 & 13 & 31.1 & 2.1 \\
    Gemma 3 12B       & 13 & 66.3 & 14 & \rkdn & 49.3 & 3.2 & 14 & \rkno & 49.4 & 1.9 & 14 & \rkno & 49.3 & 1.1 & 14 & 66.9 & 2.2 & 12 & 35.4 & 1.6 \\
    Gemma 2 27B       & 14 & 62.0 & 15 & \rkdn & 34.6 & 3.1 & 15 & \rkno & 34.4 & 2.0 & 15 & \rkno & 34.2 & 1.2 & 16 & 47.4 & 2.4 & 15 & 24.0 & 1.6 \\
    Qwen 3 32B        & 15 & 58.9 & 13 & \rkup & 51.2 & 2.8 & 13 & \rkno & 51.2 & 1.4 & 13 & \rkno & 51.2 & 1.0 & 6  & 84.6 & 1.6 & 14 & 24.4 & 1.2 \\
    Gemma 2 9B        & 16 & 52.9 & 17 & \rkdn & 30.2 & 4.3 & 17 & \rkno & 30.0 & 2.4 & 17 & \rkno & 30.2 & 1.3 & 18 & 44.6 & 2.5 & 16 & 18.4 & 2.3 \\
    Llama 3 70B       & 17 & 51.8 & 18 & \rkdn & 27.8 & 3.8 & 18 & \rkno & 27.7 & 1.9 & 18 & \rkno & 27.7 & 1.2 & 17 & 46.2 & 2.4 & 18 & 12.9 & 1.6 \\
    Gemma 3 4B        & 18 & 48.9 & 16 & \rkup & 31.6 & 3.2 & 16 & \rkno & 32.1 & 2.0 & 16 & \rkno & 32.0 & 1.3 & 15 & 52.6 & 2.4 & 17 & 15.6 & 1.7 \\
    \bottomrule
  \end{tabular}
  \caption{Model ranking and average accuracy on \icboth under various resource levels, with 3, 5, and 10 instances per problem. Sub-columns include: \faMedal\ rank, \faSort\ rank change vs.\ the preceding metric (\rkup\ rise, \rkdn\ fall, \rkno\ unchanged), \faPercent\ accuracy, $\pm$ half-width of the 95\% confidence interval. Rows are sorted by $D_O$ rank.}
  \label{tab:merged_model_rankings}
\end{table*}

\section{Experimental Setup and Results}
\subsection{Experiment Setup}
\paragraph{Models evaluated} We benchmark two broad categories of models: open and closed. For open models, we evaluated GPT-OSS series (20B and 120B) ~\cite{openai2025gptoss120bgptoss20bmodel}, Gemma 2 series (9B and 27B)\cite{team2024gemma}, Gemma 3 series (4B, 12B, and 27B)~\cite{team2025gemma}, Qwen 3 32B~\cite{yang2025qwen3}, Qwen 3.5 27B~\cite{team2026qwen3}, and Deepseek V3 0324~\cite{liu2024deepseek}. As for proprietary models, we evaluated Gemini 2.0 Flash, Gemini 2.5 Flash~\cite{comanici2025gemini}, Gemini 3 Flash, Gemini 3 Pro~\cite{gemini3models}, GPT-4.1 \cite{achiam2023gpt}, GPT-5~\cite{singh2025openai}, and Claude Sonnet 4~\cite{anthropic2025claude}. 

Each model is evaluated under a zero-shot setting across six variations within the \sym and \ic series for each language. To ensure robustness, every variation is evaluated five times using different values, and we report the mean performance across these iterations. We report the results of original data ($D_O$), \icn, \ichash and \symhash in the main paper. More results can be found in Appendix \ref{app:full_exp_results}.

\paragraph{Models configuration}
We set the decoding temperature to 0.1 for all models, except GPT-5 which only supports a temperature of 1. The maximum output length is 4096 tokens for all models. Also, to ensure the fairest possible comparison, we disable thinking mode whenever possible. For the five models that do not support non-thinking (GPT-5, Gemini 3.0 Flash, Gemini 3.0 Pro, GPT-OSS 120B, and GPT-OSS 20B), we set the thinking budget to be the lowest possible setting.

\paragraph{Prompt}
The prompt is structured to ensure the model adheres to the CoT format while including clear instructions to help numerical result capture. Our prompt suggests thinking in English since previous works show LLM reason better in English~\citep{tam2025language,qi2025models}. However, we discuss the impact of using native language in the result, with consistent conclusions (§\ref{subsec:native_reasoning}).

\begin{tcolorbox}[
    enhanced, 
    colback=MyOrange!10!white, 
    colframe=MyOrange,        
    arc=4mm,                  
    boxrule=1.5pt,  
    left=10pt, right=10pt, top=10pt, bottom=10pt,
    fontupper=\itshape,
    title=Evaluation Prompt,
    attach boxed title to top left={xshift=5mm, yshift=-\tcboxedtitleheight/2},
    boxed title style={
        colback=gray,     
        colframe=gray,    
        arc=2mm,          
        boxrule=0pt       
    },
    coltitle=white,       
    fonttitle=\bfseries\sffamily 
]
Explain your reasoning step by step in clear \textbf{English} to solve the problem. \\
Your response should end with the final numerical answer, without including units. \\
Question: \{question\}
\end{tcolorbox}

\subsection{Results}
\label{sec:results}
\subsubsection{Main results}

\autoref{tab:main_result_closed_models} and \autoref{tab:main_result_open_models} show the results of five closed and five open LLMs respectively. The models include Gemini 2.5 Flash, Gemini 3.0 Pro, Claude 4 Sonnet, GPT-4.1, GPT-5, Gemma 3 27B, Qwen 3 32B, Qwen 3.5 27B, DeepSeek V3, and GPT-OSS 120B. We highlight the main findings below.

\paragraph{LLM performance is less sensitive to name variation} Simply changing the names of people or items (i.e. \symn setting) does not necessarily hurt performance.
However, when irrelevant contexts are added (i.e. \icn), there is little drop. In general, \icn is more challenging for LRLs such as Twi, Igbo, or Yoruba than HRLs. Also, we find proprietary models to be more robust to this drop. For example, on average, Gemini 3.0 Pro accuracy on all languages dropped by $-1.8$ while open models such as DeepSeek V3 and GPT-OSS 120B dropped by $-4.7$ and $-4.0$ respectively. Full result for \symn is in Appendix ~\ref{app:full_exp_results}, omitted in main table due to space constraint.

\paragraph{Numerical variation leads to huge drop in performance} While name variation leads to a small drop, changing numbers used in the questions leads to a huge drop in performance especially when combined with irrelevant contexts. All models dropped by at least $-6.5$ points under the \ichash setting, except for the top-performing model Gemini 3.0 Pro.

\paragraph{High-resource languages are more robust to variations} 

From \autoref{tab:main_result_closed_models} and \autoref{tab:main_result_open_models}, we observe that overall models are less robust in LRL setting than HRL setting. For instance, the median accuracy drop on Twi is -12.1 and -13.5 for open and closed models, respectively. On the other hand, Chinese only exhibits -4.2 and -4.7 for open and closed models, respectively.

\paragraph{More capable recent models are more robust} 
\autoref{fig:lrl_vs_hrl_relative_decrease}  corroborate the finding that the LRL are less robust. Here, we show the comparison of the relative accuracy drop across models on HRLs and LRLs. Across all models, LRL settings have larger drop than the HRL settings, indicating that model robustness differs per language, and LRLs suffer more. We find the more recent Gemini 3.0 Flash to be more robust than Gemini 2.5 Flash, this shows newer LLMs are improving in robustness. In addition, we also find similar result comparing Qwen 3 32B and Qwen 3.5 27B.

\begin{table*}[t]
  \centering
  \footnotesize
  \setlength{\tabcolsep}{4pt}
 
  \begin{tabular}{l ccc ccc ccc ccc}
    \toprule
    Languages & \multicolumn{6}{c}{DeepSeek V3} & \multicolumn{6}{c}{Gemini 2.5 Flash} \\
    \cmidrule(lr){2-7} \cmidrule(lr){8-13}
     & \multicolumn{3}{c}{English Solve} & \multicolumn{3}{c}{Native Solve} & \multicolumn{3}{c}{English Solve} & \multicolumn{3}{c}{Native Solve} \\
    \cmidrule(lr){2-4} \cmidrule(lr){5-7} \cmidrule(lr){8-10} \cmidrule(lr){11-13}
     & $D_O$ & Sym\_\# & $\Delta$ & $D_O$ & Sym\_\# & $\Delta$ & $D_O$ & Sym\_\# & $\Delta$ & $D_O$ & Sym\_\# & $\Delta$ \\
    \midrule
    English & 98.4 & 92.7 & -5.7 & 97.6 & 91.8 & -5.8 & 96.8 & 83.1 & -13.7 & 95.2 & 83.8 & -11.4 \\
    Chinese & 92.3 & 89.9 & -2.4 & 93.2 & 90.3 & -2.9 & 89.9 & 78.0 & -11.9 & 90.0 & 79.3 & -10.7 \\
    French & 90.7 & 87.8 & -2.9 & 90.8 & 86.0 & -4.8 & 91.5 & 77.5 & -14.0 & 90.8 & 79.0 & -11.8 \\
    Japanese & 88.7 & 79.8 & -9.0 & 89.6 & 82.8 & -6.8 & 86.7 & 74.9 & -11.8 & 88.0 & 74.4 & -13.6 \\
    Swahili & 90.7 & 86.3 & -4.4 & 88.4 & 82.7 & -5.7 & 91.5 & 80.3 & -11.2 & 90.4 & 78.4 & -12.0 \\
    Amharic & 76.2 & 71.9 & -4.4 & 72.0 & 69.9 & -2.1 & 81.9 & 71.0 & -10.9 & 82.8 & 68.1 & -14.7 \\
    Igbo & 73.8 & 64.6 & -9.2 & 71.2 & 61.3 & -9.9 & 81.5 & 68.6 & -12.8 & 80.4 & 65.9 & -14.5 \\
    Yoruba & 66.1 & 57.1 & -9.0 & 62.4 & 53.7 & -8.7 & 83.9 & 69.7 & -14.2 & 81.2 & 65.9 & -15.3 \\
    Twi & 48.0 & 38.0 & -10.0 & 46.4 & 37.7 & -9.4 & 65.3 & 51.5 & -13.9 & 67.2 & 51.0 & -16.2 \\
    \midrule
    Ave. & 80.5 & 74.2 & -6.3 & 79.1 & 72.8 & -6.3 & 85.4 & 72.7 & -12.7 & 85.1 & 71.8 & -13.4 \\
    \bottomrule
  \end{tabular}
   \caption{\textbf{Comparison of native and English solve on 5-instances of \symhash with Deepseek V3 and Gemini 2.5}. We report $\Delta$(\symhash $- D_o$)}
  \label{tab:native_vs_english_solve}
\end{table*}

\paragraph{Model size do not correlate to robustness}

\autoref{fig:combined_model_scaling_comparison} shows the effect of scaling of model sizes and robustness to change in names and numbers (\icboth). There is no clear pattern across different model architectures. For the Gemma family of models, the drop in performance gets worse as the model parameters increase from 4B, 12B, and 27B (\autoref{fig:gemma_comparison}). However, for GPT-OSS, we have the opposite trend where bigger model size is more robust to the performance drop (\autoref{fig:gpt_oss_comparison}). Surprisingly, we find GPT-OSS 120B to be more robust to degradation than GPT-4.1 which may be of bigger parameter size since it is a closed model. These findings suggest that model robustness is not a direct result of model size but rather other factors, maybe such as training recipes.

\subsubsection{Reliability of Leaderboard ranking}
\label{subsubsec:leaderboard_reliability}

\paragraph{Five evaluations provide stability} Most leaderboard rankings for math reasoning are based on one instance. However, our results in \autoref{tab:merged_model_rankings} demonstrate that single-instance rankings are unstable as there are substantial shifts once models are evaluated across multiple distinct instances of \icboth. To identify a reliable evaluation protocol, we examine average accuracy, ranking volatility, and 95\% confidence interval (CI) widths across three, five, and ten instances (\textsc{Avg-3}, \textsc{Avg-5}, and \textsc{Avg-10}). While evaluating across multiple instances is essential for robust leaderboard ranking, we notice that \textsc{Avg-3} yields confidence intervals that are roughly twice as wide as those in \textsc{Avg-5}. Moreover, four of the top ten models continue to shift rank as we extend the number of evaluated instances from three to five. In contrast, \textsc{Avg-5} substantially stabilizes the leaderboard, and scaling further to \textsc{Avg-10} 
yield similar ranking to the \textsc{Avg-5} setting. These findings are interesting, since varying the questions with five instances already gave a more robust, and realistic estimation of math reasoning for the language and LLM. We therefore recommend math reasoning evaluation should use \textsc{Avg-5} setting as the default.

\paragraph{High and low resource leaderboard rankings} \autoref{tab:merged_model_rankings} shows the model rankings on five instances of \icboth for both HRLs and LRLs. Interestingly, the rankings across the two settings differ greatly. On LRLs, Gemini 3.0 Pro, 3.0 Flash, and 2.5 Flash take the top three spots. Meanwhile, DeepSeek V3 and GPT-OSS 120B trail in eighth and ninth place, respectively. Under the HRL setting, however, Gemini 2.5 Flash's ranking drops significantly to twelveth place. At the same time, DeepSeek V3 and GPT-OSS 120B rise to rank four and three, respectively. This indicates that mathematical robustness under HRL do not translate to LRL.

\section{Discussion}

\begin{table}[t]
  \centering
  \footnotesize
  \setlength{\tabcolsep}{4pt}
  \renewcommand{\arraystretch}{1.15}
  \begin{tabular}{l !{\color{black!30}\vrule} ccc !{\color{black!30}\vrule} ccc}
    \toprule
    & \multicolumn{3}{c}{\textbf{DeepSeek V3}}
    & \multicolumn{3}{c}{\textbf{Gemini 2.5 Flash}} \\
    \cmidrule(lr){2-4} \cmidrule(lr){5-7}
    \textbf{Language}
    & \faLanguage & \faBrain & \faCalculator
    & \faLanguage & \faBrain & \faCalculator \\
    \midrule
    English  &  1.2 &  5.2 & 4.0 &  0.8 &  4.8 & 14.8 \\
    Chinese  &  4.0 &  8.4 & 2.4 &  5.2 &  8.4 & 15.2 \\
    French   &  3.5 &  4.8 & 3.6 &  3.2 &  8.8 & 16.0 \\
    Japanese &  9.2 & 13.6 & 4.8 &  6.8 &  7.2 & 13.2 \\
    \midrule
    Swahili  &  7.2 & 12.8 & 3.6 &  4.8 &  7.6 & 10.0 \\
    Amharic  & 19.2 & 20.8 & 5.6 & 10.4 & 13.2 & 12.4 \\
    Igbo     & 26.0 & 23.6 & 3.6 & 22.0 & 18.0 & 10.0 \\
    Yoruba   & 39.6 & 31.2 & 6.0 & 18.4 & 15.6 & 11.2 \\
    Twi      & 62.0 & 52.4 & 3.2 & 39.2 & 30.0 & 10.8 \\
    \bottomrule
  \end{tabular}
  \caption{\textbf{Error analysis of LLM models:} DeepSeek V3 and Gemini 2.5 Flash across high- and low-resource languages via GPT-5.4 as a judge. Errors rates (\%) are categorized into \textsc{Linguistic} (\faLanguage), \textsc{Logic} (\faBrain), and \textsc{Arithmetic} (\faCalculator).}
  \label{tab:gemini_DeepSeek_error_analysis}
\end{table}

\label{sec:discussion}

While in the Result section (§\ref{sec:results}), we focused on performance degradation and ranking instability across model families, the underlying mechanisms behind this lack of robustness has not been investigated. In this section, we examine two key questions: 1) how reasoning in the target ``native'' language compared to English impacts model reasoning robustness (\autoref{subsec:native_reasoning}), and 2) how individual factors such as linguistic understanding, logical deduction, and arithmetic capability contribute to model failure (\autoref{subsec:erroranalysis}).

\subsection{Effect of Language Choice on Reasoning}
\label{subsec:native_reasoning}
One natural question is the effect of language choice on the performance of LLM when they are asked to reason in ``native'' language rather than in ``English''. While there is several evidence showing that \textit{prompting the LLM in English} tends to give worse performance especially for low-resource languages~\citep{tam2025language,qi2025models}, we need to verify if asking the model to reason in \textit{native} language reduces robustness. While all results reported are from English-solve setting, we also evaluated models under native-solve setting via the prompt below. Due to computational restraints, we limit our evaluation on DeepSeek V3 and Gemini 2.5. We selected DeepSeek-V3 and Gemini 2.5 to contrast opposing model profiles: DeepSeek-V3 offers stronger mathematical reasoning with narrower linguistic coverage, whereas Gemini 2.5 provides broader multilingual support.

\begin{tcolorbox}[
    enhanced, 
    colback=MyOrange!10!white, 
    colframe=MyOrange,        
    arc=4mm,                  
    boxrule=1.5pt,  
    left=10pt, right=10pt, top=10pt, bottom=10pt,
    fontupper=\itshape,
    title=Native Evaluation Prompt,
    attach boxed title to top left={xshift=5mm, yshift=-\tcboxedtitleheight/2},
    boxed title style={
        colback=gray,     
        colframe=gray,    
        arc=2mm,          
        boxrule=0pt       
    },
    coltitle=white,       
    fonttitle=\bfseries\sffamily 
]
Explain your reasoning step by step in the problem's \textbf{native language }to solve the problem. \\
Your response should end with the final numerical answer, without including units. \\
Question: \{question\}
\end{tcolorbox}

\paragraph{Does Native reasoning yield similar conclusions as English reasoning?}
\autoref{tab:native_vs_english_solve} compares both models under native reasoning and English reasoning on \symhash setting. As previous work pointed out, the average performance of ``Native solve'' is lower than ``English solve''. More importantly, we observe similar patterns in native-reasoning in comparison to English-reasoning, where LRLs observe a sharper drop in performance than HRLs. Interestingly, the drop in accuracy (i.e. $\Delta$(\symhash $- D_o$) is very similar with only a few exceptions: For DeepSeek, the drop in performance is smaller for Japanese and Amharic, languages with non-Latin scripts, while for Gemini 2.5 Flash, the difference of $\Delta$ is often less than $\pm2$.

\paragraph{Statistical Significance of the English-Native Reasoning Gap}

To verify whether the accuracy drop for both model accuracies due to the change of reasoning language from English to Native is statistically significant, we employ McNemar's test. Precisely, we pair each English-solve response with its corresponding Native-solve response for the same problem across all 1,240 questions per language (248 questions × 5 instances) and test the one-sided hypothesis that English is correct more often than Native on the resulting discordant pairs, at $\alpha = 0.05$.

Across both DeepSeek-V3 and Gemini-2.5-flash, the accuracy decrease from English to Native-solve is statistically significant for low-resource languages, including Swahili, Yoruba, Igbo and Amharic. In contrast, performance drops for high-resource languages like English and Chinese remain statistically insignificant. These results highlight the critical role of language familiarity for model reasoning. Interestingly, we notice that Twi's accuracy drop from English to Native-solve is not statistically significant for both models despite having the lowest absolute accuracy of any language under both models. This is most likely because the Twi dataset is hard for both English and native solve settings, and hence models perform similarly despite switching reasoning language.

\subsection{Disentangling Linguistic, Logical, and Arithmetic Failures}
\label{subsec:erroranalysis}

\autoref{tab:gemini_DeepSeek_error_analysis} compares the errors made by DeepSeek V3 and Gemini 2.5 on \symhash, classifying them into three types: linguistic misunderstandings, logical reasoning errors, and arithmetic errors. Each question can be labeled with more than one error type, since the categories are not mutually exclusive. A clear pattern emerges: linguistic errors increase substantially as we move from high-resource to low-resource languages. This trend is especially pronounced for DeepSeek V3, whose linguistic error rate rises from below 10\% on all high-resource languages to 62.0\% on Twi. Gemini 2.5 Flash shows the same trend but with consistently lower linguistic error rates, suggesting stronger multilingual comprehension under distribution shift. We also observe that these initial linguistic errors frequently propagate into logical reasoning failures, ultimately leading to incorrect answers. 

In contrast, the arithmetic error rates for both models remain relatively stable across languages, indicating that calculation abilities are largely unaffected by language shift. Overall, this suggests that the multilingual nature of MGSM-Pro adds a new layer of difficulty that directly impacts model robustness, as ultimately, a model's performance relies on both its arithmetic capability and its familiarity with the target language. 

Furthermore, we validate the reliability of the LLM judge's error classification via human verification. More details are discussed in \autoref{app:validate_llm_as_judge_error_analysis}.


\begin{figure}[t]
    \centering
    \includegraphics[width=\linewidth]{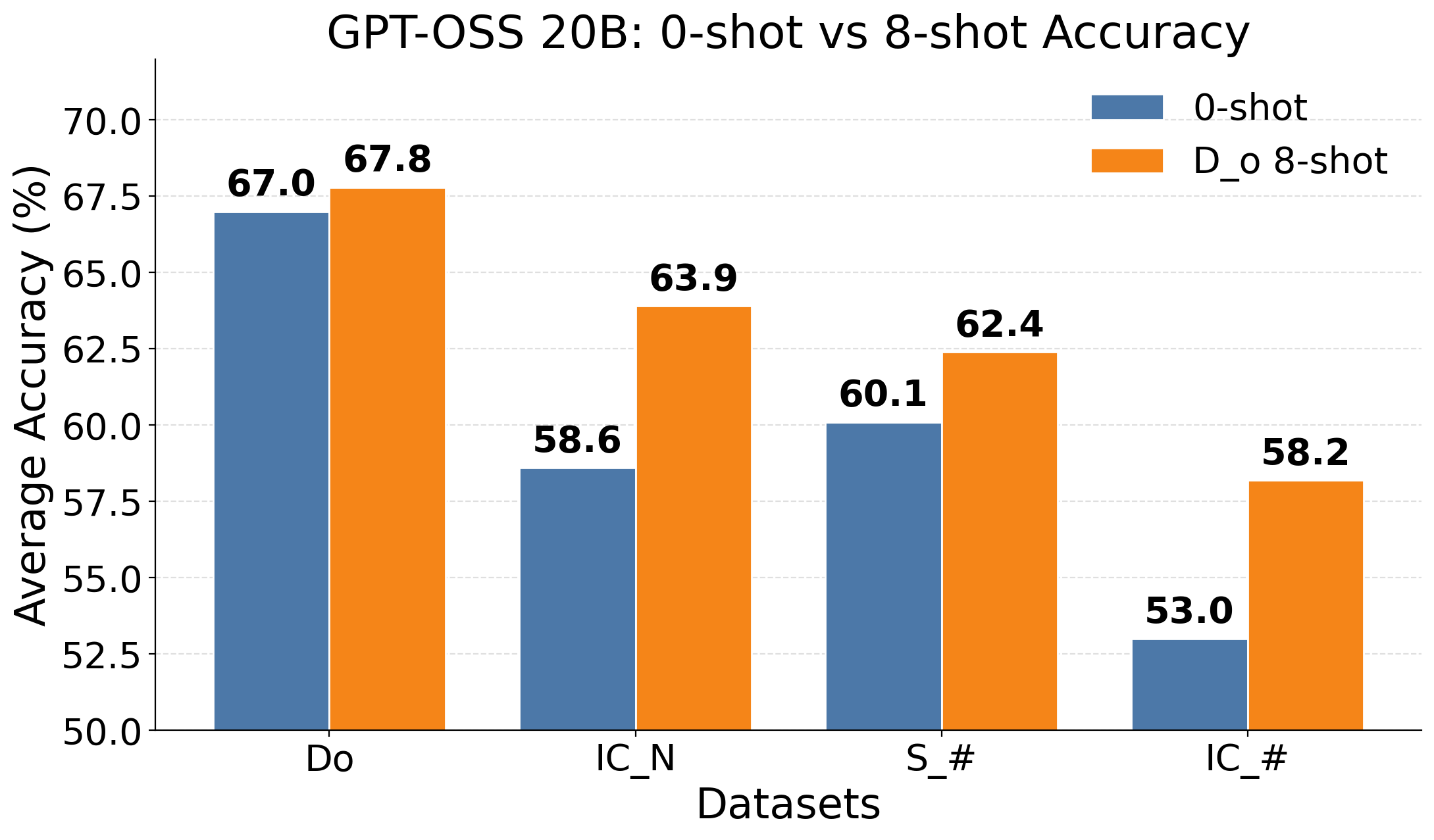}
    \caption{GPT-OSS 20B average accuracy over all languages on $D_O$, \icn, \symhash, and \ichash under 0-shot and 8-shot prompting.}
    \label{fig:gpt_oss_20b_0shot_vs_8shot}
\end{figure}

\subsection{Does few-shot prompting improve robustness?}

\autoref{fig:gpt_oss_20b_0shot_vs_8shot} compares GPT-OSS 20B under 0-shot and 8-shot prompting on $D_O$, \icn, \symhash, and \ichash, averaged across all languages. We only conducted few-shot experiments on GPT-OSS 20B because the other models often performed worse under few-shot prompting than in the zero-shot setting. For all settings, the 8-shot examples are drawn from the original training set of each respective MGSM/AfriMGSM language. We have also experimented with appending irrelevant context to these exemplars (forming "IC-few-shot" exemplars) when evaluating models under the \ic series. However, we found that the results were similar to the original few-shot setup.

On $D_O$, 8-shot prompting yielded only a modest accuracy improvement of $+0.8$. However, the impact of few-shot prompting was substantially larger for \icn, \symhash, and \ichash. The largest gains were observed in the irrelevant-context setting, where performance improved by more than $+5.0$ on average across languages. The improvement for the \symhash setting was smaller but still appears significant. These findings suggest that few-shot prompting can partially recover math reasoning performance degraded by irrelevant contexts and digit substitutions. Nevertheless, performance in both the zero-shot and few-shot settings remains below that of the original $D_O$ benchmark. This calls for a better strategy in improving mathematics robustness in multilingual settings. We provide the full results are in \ref{app:0shot_vs_8shot_results}.

\section{Conclusion}
In this paper, we investigated the robustness of LLM evaluation for math reasoning when presented with multiple instantiations of the same question by varying names, digits and adding irrelevant contexts. We developed MGSM-Pro, an extension of MGSM with five new instances per question to encourage more robust and realistic evaluation across nine typologically diverse languages. All LLMs experienced significant drops in performance, especially for low-resource languages. Furthermore, our findings reveal that reasoning robustness in high-resource languages does not transfer to low-resource settings as linguistic understanding, logical reasoning, and arithmetic capabilities are equally important for a robust multilingual math reasoner.

\section{Limitations}
Our study has a few limitations. First, our dataset covers a relatively small set of nine languages due to resource constraints. The construction process requires significant human labor to verify each of the 248 questions when converted to templates, taking almost 12 hours per language for verification. However, our approach can be easily extended to other languages, provided the resource. Expanding MGSM-Pro to other languages such as Tamil would provide a more complete picture of multilingual mathematical robustness. Moreover, our evaluation covers only 18 models because of the limited compute budget. It remains to be seen how other model families, such as Kimi, would perform. Moreover, it will be interesting to see how varying thinking mode can different thinking models can affect model robustness in MGSM-Pro.


\section*{Acknowledgment}
This research was supported in part by the Natural Sciences and Engineering Research Council (NSERC) of Canada and in part by the AI2050 program at Schmidt Sciences. We are grateful for the support of Mila’s computing resources (mila.quebec) and Digital Alliance of Canada. This work is also partially supported by Azure sponsorship credits granted by Microsoft’s AI for Good Research Lab, LLM API credits provided by
Google’s Gemini Academic Program Award and the OpenAI Researcher Access Award. We thank Haneul Yoo for her suggestions on the paper. Finally, we thank the anonymous reviewers and AC for the suggestions that greatly improved the paper.

\bibliography{custom}

\appendix

\newpage
\appendix

\section{Dataset Details}

\subsection{Dataset Statistics}

MGSM-Pro contains 9 languages, each with six dataset variants. Within each variant, we generate 5 separate subsets. Each of these subsets consists of 248 generated questions of the 250 original questions from the baseline evaluation dataset. Two questions are left out (problems 86 and 153) because it is impossible to craft 5 problem variations by changing digits. Since these two problems express all quantities exclusively as words and contain no numerical digits, they could not be used to generate the required variations as our \symhash  dataset varies only the digits from the original problem and not the numerical words for simplicity. The templates left out are the following:

\begin{tcolorbox}[
    enhanced, 
    colback=MyOrange!10!white, 
    colframe=MyOrange,        
    arc=4mm,                  
    boxrule=1.5pt,  
    left=10pt, right=10pt, top=10pt, bottom=10pt,
    title=Template 86,
    attach boxed title to top left={xshift=5mm, yshift=-\tcboxedtitleheight/2},
    boxed title style={
        colback=gray,     
        colframe=gray,    
        arc=2mm,          
        boxrule=0pt       
    },
    coltitle=white,       
    fonttitle=\bfseries\sffamily 
]
\{name\_male, Greg\} has an alarm set to ring three times a day as a reminder. When the alarm goes off, it continues to ring until \{name\_male, Greg\} turns it off. The first time it went off today, it rang four times. The second time it went off, it rang for three times as long as the first time. The third time, it rang for half as long as the second time. How many times did the alarm ring in all?
\end{tcolorbox}

\begin{tcolorbox}[
    enhanced, 
    colback=MyOrange!10!white, 
    colframe=MyOrange,        
    arc=4mm,                  
    boxrule=1.5pt,  
    left=10pt, right=10pt, top=10pt, bottom=10pt,
    title=Template 153,
    attach boxed title to top left={xshift=5mm, yshift=-\tcboxedtitleheight/2},
    boxed title style={
        colback=gray,     
        colframe=gray,    
        arc=2mm,          
        boxrule=0pt       
    },
    coltitle=white,       
    fonttitle=\bfseries\sffamily 
]
\{name\_male, Dave\} bought a large pack of french fries and ate fourteen before a hungry seagull stole the pack out of his hand. When the seagull landed, he gobbled down half the amount of french fries that \{name\_male, Dave\} ate. Then three pigeons bullied him away from the food, and each pigeon ate three fries. Later, a raccoon stole two thirds of the remaining fries. Ants carried off a final french fry, leaving five behind. How many french fries were in the pack when \{name\_male, Dave\} bought it?
\end{tcolorbox}

Now, to determine the number of questions per language and dataset variant combination (\symhash, \symboth, \symn, \ichash, \icboth, \icn), we calculate a total of 1,240 questions (5 instances * 248 questions). To find the total questions per language, we multiply this by the six variants, resulting in 7,440 questions per language (1,240 questions * 6 variants). Finally, across all 9 languages, the complete MGSM-Pro dataset comprises of 66,960 generated instances (7,440 questions * 9 languages).

\subsection{Language Details}
\label{app:language_details}
The resource levels and language families of the nine languages in MGSM-Pro are shown in Table \ref{tab:language_specs}. 

\begin{table}[h]
\centering
\small
\setlength{\tabcolsep}{3pt}
\begin{tabular}{llll}
\hline
\textbf{Language} & \textbf{Code} & \textbf{Language Family} & \textbf{Joshi Class} \\
\hline
English  & eng\_Latn & Indo-European & Class 5 \\
Chinese  & zho\_Hans & Sino-Tibetan  & Class 5 \\
French   & fra\_Latn & Indo-European & Class 5 \\
Japanese & jpn\_Jpan & Japonic       & Class 5 \\
Swahili  & swh\_Latn & Niger-Congo   & Class 2 \\
Amharic  & amh\_Ethi & Afro-Asiatic  & Class 2 \\
Igbo     & ibo\_Latn & Niger-Congo   & Class 1 \\
Yoruba   & yor\_Latn & Niger-Congo   & Class 2 \\
Twi      & twi\_Latn & Niger-Congo   & Class 1 \\
\hline
\end{tabular}
\caption{Selected languages categorized by ISO code, linguistic family, and resource availability (Joshi Class).}
\label{tab:language_specs}
\end{table}

\subsection{Name Categories}
\label{app:name_categories}
To localize the templates, annotators for each language compiled a list of eight noun categories (e.g., male names, city names) and provided 20 culturally relevant examples per category. For instance, under the city category of the Swahili list, a city that mainly speaks Swahili, "Dar es Salaam” is included. Table \ref{tab:name_domain} illustrates the domains and specific name types extracted from the original problems.

\begin{table}[h]
\centering
\small
\setlength{\tabcolsep}{3pt}
\begin{tabular}{ll}
\hline
\textbf{Domain} & \textbf{Name Types} \\
\hline
People  & Male name, Female name, Family name \\
Places  & City name, Mountain name \\
Pet   & Dragon name, Dinosaur name, Cat name \\
\hline
\end{tabular}
\caption{Grouped name variables categorized by domain}
\label{tab:name_domain}
\end{table}

\subsection{IC Template Construction}
Each question in the MGSM-Pro dataset is paired with a corresponding IC sentence template. The curation of IC sentence template follows the methodology of \cite{shi-etal-2023},  where we ensure
that irrelevant sentences have: 1) some related con-
nection with the problem and 2) use names found
in the question. An example of a problem template with its corresponding IC template is shown below.

\begin{tcolorbox}[
    enhanced, 
    colback=MyOrange!10!white, 
    colframe=MyOrange,        
    arc=4mm,                  
    boxrule=1.5pt,  
    left=10pt, right=10pt, top=10pt, bottom=10pt,
    title=Problem Template,
    attach boxed title to top left={xshift=5mm, yshift=-\tcboxedtitleheight/2},
    boxed title style={
        colback=gray,     
        colframe=gray,    
        arc=2mm,          
        boxrule=0pt       
    },
    coltitle=white,       
    fonttitle=\bfseries\sffamily 
]
\{\colorbox{yellow}{name\_female}, Janet\}'s ducks lay \{\colorbox{green!50}{num\_digit\_1}, 16\} eggs per day. She eats three for breakfast every morning and bakes muffins for her friends every day with four. She sells the remainder at the farmers' market daily for \$\{\colorbox{green!50}{num\_digit\_2}, 2\} per fresh duck egg. How much in dollars does she make every day at the farmers' market?
\end{tcolorbox}

\begin{tcolorbox}[
    enhanced, 
    colback=MyOrange!10!white, 
    colframe=MyOrange,        
    arc=4mm,                  
    boxrule=1.5pt,  
    left=10pt, right=10pt, top=10pt, bottom=10pt,
    title=Corresponding IC Template,
    attach boxed title to top left={xshift=5mm, yshift=-\tcboxedtitleheight/2},
    boxed title style={
        colback=gray,     
        colframe=gray,    
        arc=2mm,          
        boxrule=0pt       
    },
    coltitle=white,       
    fonttitle=\bfseries\sffamily 
]
\{\colorbox{yellow}{name\_female}, Janet\}'s uncle brings 41 apples to her every week.
\end{tcolorbox}

\section{Annotation Protocol}
\label{app:annotation_protocol}

All multilingual templates are first translated from the English template via Gemini 2.0 Flash. Afterwards, they are reviewed by a native speaker. Then, all annotated templates were subjected to an automated verification against the original English source template. Any templates that fail the automated check was flagged for another round of annotation with the native annotator and main author. We discuss each component of the process in more detail below. 

\subsection{LLM Template Construction Prompt}
\label{app:template_construction_prompts}

The prompt used to translate multilingual template is as follow:

\begin{tcolorbox}[
    enhanced, 
    colback=MyOrange!10!white, 
    colframe=MyOrange,        
    arc=4mm,                  
    boxrule=1.5pt,  
    left=10pt, right=10pt, top=10pt, bottom=10pt,
    fontupper=\itshape,
    title=LLM Template Translation Prompt,
    attach boxed title to top left={xshift=5mm, yshift=-\tcboxedtitleheight/2},
    boxed title style={
        colback=gray,     
        colframe=gray,    
        arc=2mm,          
        boxrule=0pt       
    },
    coltitle=white,       
    fonttitle=\bfseries\sffamily 
]
Your task is to convert an English template into a native-language template, preserving all placeholder formats.
Do not change the ordering of words of the output sentence, just label them with brackets as shown below: 
\\
Input:\\
- English template (with placeholders)\\
- Native sentence (no placeholders)
\\
Output:\\
- Native sentence with the same placeholders, matching values and positions.
\\
Placeholder formats:\\
- Names: \{name\_male, xxx\}, \{name\_female, xxx\}\\
- Digits: \{num\_digit, xxx\}
\\
Guidelines:\\
1. Tag all placeholders from English in the native sentence.\\
2. Names may differ across languages (e.g. James → Jacques) — match by position.\\
3. Always tag the first word if it’s a person name.\\
4. Do not reword the native sentence in any way, you should just be inserting the variable names and brackets \\

Input: \{english template\}\\
Native: \{native question\}\\
Output: 
\end{tcolorbox}

\subsection{Template Correction Process with Annotators}
To ensure template quality, the lead author conducted on-boarding sessions with each annotator to clarify the guidelines, answer questions, and resolve any early annotations that deviated from the criteria. The MGSM-Pro annotation process is divided into two main tasks: 1) correcting native templates and 2) providing native names. The instructions for both tasks are shown below, respectively. 

The proportion of questions requiring correction varies substantially across languages, as shown in Table~\ref{tab:correction_rate}. Higher-resource languages such as Chinese, French, and Japanese required edits to fewer than 10\% of templates, whereas the lower resource languages (Swahili, Amharic, Igbo, Yoruba, and Twi) required corrections to roughly two-thirds or more of their templates.

\begin{table}[h]
\centering
\begin{tabular}{lc}
\toprule
\textbf{Language} & \textbf{Template Correction Rate (\%)} \\
\midrule
English  & --   \\
Chinese  & 4.4  \\
French   & 7.6  \\
Japanese & 9.2  \\
Swahili  & 72.4 \\
Amharic  & 67.2 \\
Igbo     & 71.2 \\
Yoruba   & 70.8 \\
Twi      & 70.4 \\
\bottomrule
\end{tabular}
\caption{Percentage of templates requiring correction by native annotators, per language. English is the source language and therefore not applicable.}
\label{tab:correction_rate}
\end{table}

\subsubsection{Annotator Template Correction Instruction}
\begin{lstlisting}[
    basicstyle=\ttfamily, 
    breaklines=true, 
    postbreak=\mbox{\textcolor{purple}{$\hookrightarrow$}\space}
]
For each problem template correction, 3 items will be provided for you to use.

1. English Template
This is the gold template. You should make sure the native language template is as similar to the english template as possible.

2. Original Native Question
This is the original native question in the dataset. You should use this as a reference alongside the English template to judge if the Native language template is correct.

3. Native Language Template
This is a machine-created native language template. It could very likely contain errors. This is the template that you will judge if it is correct or not. 

Below are the five critierias the native language template must achieve in order to be considered as correct. 

1. Native Language Templates will need to contain the original question. I.E. the wording of the native template should not change from the native question, the template should only be adding in the variable names. If this is not the case, you should ignore the Native Language Template and please provide the new annotated template inside the correction column 

2. No missing variable annotation. I.E. all names or digits tagged in English template is tagged in the native language template. You should add the corresponding {type, value} annotation around the target language word or number.

3. No extra annotation. I.E. there is no extra variables annotated in the translation but was not in the English template. You should remove any { } markers around words or numbers that were not annotated in English.

4. No incorrect bracket {} span. I.E. the annotated span is not too long or too short. You should adjust the braces so they exactly enclose the intended word or number, matching the English span.

\end{lstlisting}

\subsubsection{Annotator Native Name Annotation Instruction}

\begin{lstlisting}[
    basicstyle=\ttfamily, 
    breaklines=true, 
    postbreak=\mbox{\textcolor{purple}{$\hookrightarrow$}\space}
]
You will be given eight types of name. 

You will need to provide 10 names to these name types that fit into your native language. The name should be relevant to your specific language and not English names. However, if there are more than one and less than 10 unique names for a specific name cateogry, it is fine to provide less. Moreover, if there are no native names for a specific name category, you can provide english substitute. 

The list of name types are as follows:

Male name, Female name, Family name, City name, Mountain name, Dragon name, Dinosaur name, Cat name

\end{lstlisting}

\subsection{Template Automated Alignment Checks}

After the annotators first round of correction on the native templates, each template will go through a round of automated alignment checks to ensure template quality. Specifically, three criteria were evaluated:

\begin{enumerate}
    \item \textbf{Variable frequency}: Variables such as \verb|{name_...}| and \verb|{num_...}| must appear the same number of times in the native template as they do in the English template.
    \item \textbf{Syntax accuracy}: There must be no typo errors in the variable names of the native template(e.g. misspelling of \verb|{nam_...}|).
    \item \textbf{Variable consistency}: No new variables should be introduced in the native template that did not exist in the English source.
\end{enumerate}

Any template that fails the automated checks is flagged for the next round of annotation with the native annotator and main author. 

\section{Model Evaluation Configuration}
All models have a decoding temperature of 0.1, and the only exception is GPT-5 where it only takes in a decoding temperature of 1.0. All models have a max token count of 4096. Prompt format is consistent across all models, where system prompts are not used and only a single user prompt is used per request.

\section{Validate LLM-as-judge Error Analysis}
\label{app:validate_llm_as_judge_error_analysis}
To validate the reliability of the LLM judge's error classifications, we conducted human verification on two languages: Chinese (high-resource) and Yoruba (low-resource). This verification process evaluated the outputs of Gemini 2.5, with two human annotators assigned per language. Each annotator independently examined all errors made by Gemini 2.5 within a single \symhash dataset of 248 problems to determine if they agree with the decisions made by the GPT-5.4 judge. The results show a very strong human-model agreement, exceeding 96\% across all evaluated cases, refer to \autoref{tab:agreement-zh} and \autoref{tab:agreement-yo}. 

\begin{table}[t]
\centering
\small
\setlength{\tabcolsep}{4pt}
\begin{tabular}{@{}lcccc@{}}
\toprule
 & \textbf{Misjudg.} & \textbf{Lang.} & \textbf{Logic} & \textbf{Arith.} \\
\midrule
Human 1 & 98.25 & 96.49 & 96.49 & 100.0 \\
Human 2 & 100.0 & 96.49 & 98.25 & 100.0 \\
\bottomrule
\end{tabular}
\caption{Human--model agreement rates (\%) on error classification for Chinese.
Misjudg.\ = Misjudgment, Lang.\ = Language Understanding,
Logic = Logical Reasoning, Arith.\ = Arithmetic Calculation.}
\label{tab:agreement-zh}
\end{table}

\begin{table}[t]
\centering
\small
\setlength{\tabcolsep}{4pt}
\begin{tabular}{@{}lcccc@{}}
\toprule
 & \textbf{Misjudg.} & \textbf{Lang.} & \textbf{Logic} & \textbf{Arith.} \\
\midrule
Human 1 & 98.68 & 96.05 & 92.11 & 98.68 \\
Human 2 & 100.0 & 100.0 & 90.67 & 98.67 \\
\bottomrule
\end{tabular}
\caption{Human--model agreement rates (\%) on error classification for Yoruba.
Column abbreviations as in Table~\ref{tab:agreement-zh}.}
\label{tab:agreement-yo}
\end{table}

\subsection{0-Shot VS 8-Shot Results}
\label{app:0shot_vs_8shot_results} 

\autoref{tab:multishot_gpt_oss_20b} illustrates the 0-Shot vs 8-Shot performance of GPT-OSS-20B on 4 different dataset settings. 

\begin{table}[t]
\centering
\small
\setlength{\tabcolsep}{3pt}
\renewcommand{\arraystretch}{1.1}
\begin{tabular}{l cc cc cc cc}
\toprule
\textbf{Lang.} & \multicolumn{2}{c}{\textbf{Do}} & \multicolumn{2}{c}{\textbf{Sym\#}} & \multicolumn{2}{c}{\textbf{ICN}} & \multicolumn{2}{c}{\textbf{IC\#}} \\
\cmidrule(lr){2-3} \cmidrule(lr){4-5} \cmidrule(lr){6-7} \cmidrule(lr){8-9}
& 0-S & 8-S & 0-S & 8-S & 0-S & 8-S & 0-S & 8-S \\
\midrule
English  & 95.6 & 97.6 & 88.9 & 91.1 & 87.7 & 93.0 & 80.6 & 86.8 \\
Chinese  & 89.1 & 88.7 & 84.4 & 86.6 & 86.6 & 88.1 & 81.9 & 84.0 \\
French   & 87.9 & 87.5 & 84.1 & 84.6 & 85.6 & 87.3 & 81.9 & 82.7 \\
Japanese & 85.1 & 85.5 & 81.3 & 80.8 & 79.0 & 82.0 & 77.2 & 79.8 \\
Swahili  & 74.6 & 73.0 & 66.2 & 68.7 & 61.5 & 68.5 & 53.9 & 61.5 \\
Amharic  & 39.5 & 44.0 & 32.4 & 36.8 & 28.1 & 36.9 & 22.4 & 32.2 \\
Igbo     & 56.0 & 60.1 & 47.2 & 49.9 & 47.1 & 55.4 & 37.3 & 44.8 \\
Yoruba   & 52.0 & 49.6 & 39.8 & 41.5 & 38.6 & 44.4 & 30.3 & 35.8 \\
Twi      & 23.0 & 24.2 & 16.6 & 21.2 & 13.4 & 19.8 & 11.6 & 16.4 \\
\midrule
Ave.     & 67.0 & 67.8 & 60.1 & 62.4 & 58.6 & 63.9 & 53.0 & 58.2 \\
\bottomrule
\end{tabular}
\caption{0-Shot(0-S) and 8-Shot(8-S) performance across Do, \symhash, \icn, and \ichash for various languages for GPT-OSS-20B}
\label{tab:multishot_gpt_oss_20b}
\end{table}

\subsection{Full Experiment Results}
\label{app:full_exp_results} 

We report \icn, \symhash, and \ichash in \autoref{tab:main_result_closed_models} and \autoref{tab:main_result_open_models} because these three capture the overall trends. The appendix results follow a consistent pattern: changing only names barely affects performance, while changing numbers or adding irrelevant context leads to much larger drops. Changing both names and numbers (\symboth / \icboth) yields results close to changing numbers alone (\symhash / \ichash), so \icn, \symhash, and \ichash are sufficient to show where the degradation comes from.

Moreover, the 10 models shown in \autoref{tab:main_result_closed_models} and \autoref{tab:main_result_open_models} were chosen because they are the 10 high-performing closed-source and open-source models. The 8 weaker models display similar trends with a steeper performance drop across the MGSM-Pro dataset. The full results are shown in \autoref{tab:main_result_one}, \autoref{tab:main_result_two}, \autoref{tab:main_result_three}, \autoref{tab:main_result_four}, \autoref{tab:main_result_five}, \autoref{tab:main_result_six}.

\begin{table*}[t] 
\centering
\resizebox{\textwidth}{!}{
\setlength{\tabcolsep}{3pt} 
\renewcommand{\arraystretch}{1.1}
\begin{tabular}{l lllllll | lllllll | lllllll}
\toprule
& \multicolumn{7}{c}{\textbf{Gemini 2.0 Flash}} & \multicolumn{7}{c}{\textbf{Gemini 2.5 Flash}} & \multicolumn{7}{c}{\textbf{Gemini 3 Flash}} \\
\cmidrule(lr){2-8} \cmidrule(lr){9-15} \cmidrule(lr){16-22} 
\textbf{Language} & $D_O$ & \symn & \icn & \symhash & \ichash & \symboth & \icboth & $D_O$ & \symn & \icn & \symhash & \ichash & \symboth & \icboth & $D_O$ & \symn & \icn & \symhash & \ichash & \symboth & \icboth \\
\midrule
English  & 96.0 & 94.1 & 94.0 & 85.2 & 84.3 & 84.5 & 81.4 & 96.8 & 95.0 & 94.6 & 83.1 & 81.0 & 80.6 & 80.1 & 98.8 & 97.3 & 95.9 & 95.3 & 94.0 & 94.9 & 93.3 \\
        
Chinese  & 86.7 & 89.4 & 86.5 & 80.0 & 77.9 & 80.7 & 76.5 & 89.9 & 90.7 & 89.0 & 78.0 & 79.0 & 78.3 & 76.2 & 93.1 & 92.3 & 91.5 & 91.7 & 91.4 & 91.3 & 90.2 \\
        
French   & 90.7 & 86.7 & 87.2 & 77.2 & 76.5 & 78.5 & 76.7 & 91.5 & 87.3 & 87.1 & 77.5 & 73.8 & 75.8 & 73.5 & 92.3 & 89.7 & 89.2 & 88.1 & 87.6 & 87.1 & 86.5 \\
        
Japanese & 84.3 & 84.4 & 82.1 & 75.9 & 72.4 & 74.4 & 70.4 & 86.7 & 85.0 & 83.9 & 74.9 & 73.1 & 73.2 & 71.7 & 90.7 & 89.5 & 88.5 & 88.9 & 88.4 & 88.6 & 88.0 \\
        
Swahili  & 91.9 & 90.8 & 87.6 & 79.0 & 77.9 & 78.1 & 77.3 & 91.5 & 90.7 & 89.9 & 80.3 & 78.7 & 78.1 & 77.7 & 96.8 & 94.0 & 93.0 & 91.6 & 90.8 & 90.5 & 89.4 \\
        
Amharic  & 80.2 & 80.2 & 80.6 & 67.3 & 68.0 & 66.9 & 66.5 & 81.9 & 83.9 & 81.9 & 71.0 & 70.2 & 71.0 & 69.3 & 86.3 & 85.7 & 84.4 & 81.9 & 81.5 & 82.4 & 80.2 \\
        
Igbo     & 78.6 & 80.0 & 74.2 & 63.4 & 59.8 & 65.6 & 61.0 & 81.5 & 80.3 & 79.7 & 68.6 & 66.6 & 70.2 & 68.0 & 86.7 & 87.4 & 84.2 & 79.9 & 78.0 & 80.1 & 78.0 \\
        
Yoruba   & 79.4 & 77.9 & 71.6 & 64.4 & 60.2 & 63.5 & 59.8 & 83.9 & 83.5 & 79.4 & 69.7 & 68.1 & 69.4 & 65.8 & 83.9 & 84.0 & 82.2 & 79.9 & 76.4 & 78.7 & 74.8 \\
        
Twi      & 62.5 & 63.5 & 58.1 & 50.4 & 47.3 & 51.0 & 45.3 & 65.3 & 66.3 & 61.9 & 51.5 & 49.6 & 52.7 & 48.3 & 73.8 & 72.5 & 71.1 & 66.1 & 61.0 & 65.1 & 60.1 \\

\midrule
Average  & 83.4 & 83.0 & 80.2 & 71.4 & 69.4 & 71.5 & 68.3 & 85.4 & 84.7 & 83.0 & 72.7 & 71.1 & 72.1 & 70.1 & 89.2 & 88.0 & 86.7 & 84.8 & 83.2 & 84.3 & 82.3 \\
\bottomrule
\end{tabular}%
}
\caption{Different models' accuracy across different dataset variations ($D_O$, \symn, \icn, \symhash, \ichash, \symboth, \icboth) for each language}
\label{tab:main_result_one}
\end{table*}

\begin{table*}[t] 
\centering
\resizebox{\textwidth}{!}{
\setlength{\tabcolsep}{3pt} 
\renewcommand{\arraystretch}{1.1}
\begin{tabular}{l ccccccc | ccccccc | ccccccc}
\toprule
& \multicolumn{7}{c}{\textbf{Gemini 3.0 Pro}} & \multicolumn{7}{c}{\textbf{GPT 4.1}} & \multicolumn{7}{c}{\textbf{GPT 5}} \\
\cmidrule(lr){2-8} \cmidrule(lr){9-15} \cmidrule(lr){16-22} 
\textbf{Language} & $D_O$ & \symn & \icn & \symhash & \ichash & \symboth & \icboth & $D_O$ & \symn & \icn & \symhash & \ichash & \symboth & \icboth & $D_O$ & \symn & \icn & \symhash & \ichash & \symboth & \icboth \\
\midrule
English  & 98.0 & 96.3 & 96.2 & 94.4 & 93.2 & 93.5 & 92.8 & 96.4 & 94.7 & 91.7 & 81.5 & 79.6 & 81.1 & 79.9 & 96.8 & - & 93.3 & 92.8 & - & - & 87.6 \\
        
Chinese  & 93.5 & 93.5 & 93.1 & 93.2 & 93.5 & 92.5 & 92.3 & 89.9 & 91.4 & 90.6 & 78.6 & 76.8 & 78.3 & 76.4 & 92.3 & - & 91.9 & 89.3 & - & - & 87.1 \\
        
French   & 90.7 & 89.3 & 89.7 & 88.6 & 87.2 & 88.5 & 86.0 & 88.7 & 86.7 & 85.5 & 75.9 & 74.5 & 76.0 & 74.0 & 89.9 & - & 88.4 & 85.4 & - & - & 83.0 \\
        
Japanese & 90.7 & 89.0 & 89.4 & 87.6 & 88.0 & 88.0 & 87.2 & 87.1 & 85.8 & 83.6 & 74.8 & 74.1 & 73.5 & 73.2 & 90.7 & - & 83.6 & 83.6 & - & - & 81.1 \\
        
Swahili  & 97.6 & 94.5 & 93.9 & 93.0 & 92.4 & 91.3 & 91.2 & 91.5 & 90.0 & 89.0 & 79.8 & 77.5 & 77.5 & 77.3 & 90.7 & - & 87.7 & 87.7 & - & - & 87.9 \\
        
Amharic  & 87.5 & 86.8 & 85.4 & 83.2 & 83.0 & 82.8 & 82.2 & 68.1 & 69.4 & 68.2 & 53.0 & 51.2 & 52.7 & 51.2 & 71.8 & - & 67.6 & 67.6 & - & - & 65.9 \\
        
Igbo     & 89.9 & 88.8 & 87.5 & 82.1 & 81.2 & 82.2 & 81.1 & 79.0 & 77.2 & 71.3 & 58.5 & 54.0 & 59.4 & 55.5 & 79.8 & - & 70.2 & 70.2 & - & - & 65.8 \\
        
Yoruba   & 86.3 & 86.5 & 85.7 & 83.1 & 80.9 & 81.0 & 77.8 & 72.6 & 73.2 & 69.6 & 59.8 & 55.6 & 59.0 & 55.2 & 74.6 & - & 67.7 & 67.7 & - & - & 65.4 \\
        
Twi      & 77.8 & 77.1 & 74.9 & 70.6 & 69.1 & 69.8 & 66.0 & 41.9 & 43.5 & 36.9 & 31.5 & 28.4 & 31.7 & 26.2 & 46.8 & - & 38.5 & 38.5 & - & - & 37.5 \\

\midrule
Average  & 90.2 & 89.1 & 88.4 & 86.2 & 85.4 & 85.5 & 84.1 & 79.5 & 79.1 & 76.3 & 65.9 & 63.5 & 65.5 & 63.2 & 81.5 & - & 79.5 & 75.9 & - & - & 73.5 \\
\bottomrule
\end{tabular}%
}
\caption{Different models' accuracy across different dataset variations ($D_O$, \symn, \icn, \symhash, \ichash, \symboth, \icboth) for each language}
\label{tab:main_result_two}
\end{table*}

\begin{table*}[t] 
\centering
\resizebox{\textwidth}{!}{
\setlength{\tabcolsep}{3pt} 
\renewcommand{\arraystretch}{1.1}
\begin{tabular}{l ccccccc | ccccccc | ccccccc}
\toprule
& \multicolumn{7}{c}{\textbf{GPT-OSS 20 B}} & \multicolumn{7}{c}{\textbf{GPT-OSS 120B}} & \multicolumn{7}{c}{\textbf{DeepSeek V3}} \\
\cmidrule(lr){2-8} \cmidrule(lr){9-15} \cmidrule(lr){16-22} 
\textbf{Language} & $D_O$ & \symn & \icn & \symhash & \ichash & \symboth & \icboth & $D_O$ & \symn & \icn & \symhash & \ichash & \symboth & \icboth & $D_O$ & \symn & \icn & \symhash & \ichash & \symboth & \icboth \\
\midrule
English  & 95.6 & 94.9 & 87.7 & 88.9 & 80.6 & 88.1 & 78.8 & 96.4 & 96.2 & 93.9 & 93.5 & 92.1 & 92.0 & 90.6 & 98.4 & 96.6 & 94.3 & 92.7 & 89.7 & 91.9 & 89.5 \\
        
Chinese  & 89.1 & 89.0 & 86.6 & 84.4 & 81.9 & 84.0 & 81.3 & 91.5 & 92.3 & 90.6 & 90.2 & 89.7 & 90.4 & 88.1 & 92.3 & 93.2 & 91.4 & 89.9 & 88.1 & 89.1 & 87.2 \\
        
French   & 87.9 & 87.3 & 85.6 & 84.1 & 81.9 & 83.3 & 79.9 & 90.7 & 88.6 & 88.7 & 86.5 & 86.5 & 86.5 & 83.9 & 90.7 & 89.9 & 88.7 & 87.8 & 85.7 & 86.6 & 85.7 \\
        
Japanese & 85.1 & 83.2 & 79.0 & 81.3 & 77.2 & 81.0 & 75.7 & 89.1 & 86.6 & 85.2 & 85.2 & 83.9 & 84.5 & 82.9 & 88.7 & 83.9 & 83.0 & 79.8 & 79.8 & 80.8 & 79.8 \\
        
Swahili  & 74.6 & 74.9 & 61.5 & 66.2 & 53.9 & 64.1 & 54.7 & 84.7 & 84.7 & 82.2 & 81.5 & 79.6 & 81.4 & 79.5 & 90.7 & 90.5 & 89.6 & 86.3 & 84.0 & 85.9 & 85.2 \\
        
Amharic  & 39.5 & 40.4 & 28.1 & 32.4 & 22.4 & 32.1 & 21.4 & 60.9 & 64.5 & 57.7 & 58.5 & 53.7 & 58.9 & 52.3 & 76.2 & 77.2 & 73.5 & 71.9 & 66.0 & 71.9 & 66.9 \\
        
Igbo     & 56.0 & 60.1 & 47.1 & 47.2 & 37.3 & 46.9 & 37.9 & 77.4 & 78.1 & 73.3 & 67.7 & 64.7 & 68.1 & 63.4 & 73.8 & 74.7 & 67.8 & 64.6 & 58.0 & 64.2 & 58.7 \\
        
Yoruba   & 52.0 & 50.6 & 38.6 & 39.8 & 30.3 & 39.1 & 30.2 & 76.2 & 71.7 & 66.9 & 66.1 & 61.3 & 65.1 & 59.6 & 66.1 & 65.9 & 58.0 & 57.1 & 52.7 & 58.1 & 51.9 \\
        
Twi      & 23.0 & 21.3 & 13.4 & 16.6 & 11.6 & 16.1 & 11.2 & 44.8 & 46.9 & 37.7 & 39.0 & 31.5 & 37.9 & 33.1 & 48.0 & 43.7 & 36.9 & 38.0 & 28.1 & 39.4 & 29.4 \\

\midrule
Average  & 67.0 & 66.8 & 58.6 & 60.1 & 53.0 & 59.4 & 52.3 & 79.1 & 78.9 & 75.1 & 74.2 & 71.4 & 73.9 & 70.4 & 80.6 & 79.5 & 75.9 & 74.2 & 70.3 & 74.2 & 70.5 \\
\bottomrule
\end{tabular}%
    }
\caption{Different models' accuracy across different dataset variations ($D_O$, \symn, \icn, \symhash, \ichash, \symboth, \icboth) for each language}
\label{tab:main_result_three}
\end{table*}

\begin{table*}[t] 
\centering
\resizebox{\textwidth}{!}{
\setlength{\tabcolsep}{3pt} 
\renewcommand{\arraystretch}{1.1}
\begin{tabular}{l lllllll | lllllll | lllllll}
\toprule
& \multicolumn{7}{c}{\textbf{Gemma 3 4B}} & \multicolumn{7}{c}{\textbf{Gemma 3 12B}} & \multicolumn{7}{c}{\textbf{Gemma 3 27B}} \\
\cmidrule(lr){2-8} \cmidrule(lr){9-15} \cmidrule(lr){16-22} 
\textbf{Language} & $D_O$ & \symn & \icn & \symhash & \ichash & \symboth & \icboth & $D_O$ & \symn & \icn & \symhash & \ichash & \symboth & \icboth & $D_O$ & \symn & \icn & \symhash & \ichash & \symboth & \icboth \\
\midrule
English  & 85.5 & 85.6 & 80.0 & 68.1 & 64.0 & 66.5 & 61.3 & 92.7 & 92.8 & 92.7 & 76.9 & 77.3 & 75.8 & 76.5 & 95.6 & 94.4 & 93.3 & 81.6 & 77.6 & 80.2 & 77.6 \\
        
Chinese  & 76.2 & 78.6 & 67.8 & 63.5 & 54.0 & 60.2 & 53.1 & 86.7 & 88.0 & 84.9 & 70.9 & 65.6 & 70.6 & 66.0 & 87.9 & 89.7 & 85.5 & 75.2 & 72.3 & 76.4 & 72.1 \\
        
French   & 79.0 & 78.2 & 69.3 & 59.0 & 53.8 & 60.3 & 52.7 & 87.9 & 85.9 & 81.8 & 71.7 & 66.2 & 70.2 & 65.4 & 89.1 & 87.4 & 84.5 & 73.4 & 69.9 & 74.4 & 69.0 \\
        
Japanese & 70.2 & 67.3 & 58.1 & 51.9 & 45.2 & 52.1 & 43.4 & 83.5 & 81.2 & 79.4 & 66.1 & 60.2 & 65.3 & 59.8 & 85.1 & 83.1 & 79.4 & 71.1 & 64.6 & 70.5 & 64.8 \\
        
Swahili  & 59.7 & 64.4 & 55.6 & 48.7 & 40.6 & 46.6 & 39.0 & 81.5 & 86.5 & 81.4 & 65.6 & 61.1 & 65.6 & 61.7 & 89.1 & 87.7 & 85.4 & 72.2 & 71.2 & 72.3 & 70.1 \\
        
Amharic  & 40.7 & 42.9 & 37.7 & 32.7 & 26.0 & 32.3 & 27.2 & 68.1 & 69.2 & 67.9 & 56.6 & 52.3 & 55.6 & 54.5 & 71.4 & 71.9 & 70.2 & 57.4 & 55.1 & 56.8 & 53.8 \\
        
Igbo     & 16.5 & 15.0 & 13.5 & 10.6 & 7.5  & 10.2 & 7.6  & 55.6 & 53.2 & 44.4 & 38.1 & 33.0 & 38.9 & 34.8 & 64.1 & 65.0 & 54.0 & 50.0 & 41.3 & 51.9 & 42.8 \\
        
Yoruba   & 7.7  & 9.0  & 4.9  & 4.8  & 3.5  & 5.2  & 3.3  & 32.7 & 34.4 & 26.8 & 24.9 & 20.2 & 26.0 & 21.7 & 44.4 & 49.1 & 40.5 & 38.2 & 31.4 & 38.1 & 31.9 \\
        
Twi      & 4.4  & 3.5  & 1.9  & 1.8  & 0.9  & 2.1  & 1.0  & 7.7  & 8.2  & 6.2  & 7.0  & 4.3  & 5.6  & 4.4  & 18.1 & 17.1 & 10.4 & 13.4 & 9.3  & 14.0 & 9.4  \\

\midrule
Average  & 48.9 & 49.4 & 43.2 & 37.9 & 32.8 & 37.3 & 32.1 & 66.3 & 66.6 & 62.8 & 53.1 & 48.9 & 52.6 & 49.4 & 71.6 & 71.7 & 67.0 & 59.2 & 54.7 & 59.4 & 54.6 \\
\bottomrule
\end{tabular}%
}
\caption{Different models' accuracy across different dataset variations ($D_O$, \symn, \icn, \symhash, \ichash, \symboth, \icboth) for each language}
\label{tab:main_result_four}
\end{table*}

\begin{table*}[t] 
\centering
\resizebox{\textwidth}{!}{
\setlength{\tabcolsep}{3pt} 
\renewcommand{\arraystretch}{1.1}
\begin{tabular}{l ccccccc | ccccccc | ccccccc}
\toprule
& \multicolumn{7}{c}{\textbf{Claude 4}} & \multicolumn{7}{c}{\textbf{Llama 3 70B}} & \multicolumn{7}{c}{\textbf{Gemma 2 27B}} \\
\cmidrule(lr){2-8} \cmidrule(lr){9-15} \cmidrule(lr){16-22} 
\textbf{Language} & $D_O$ & \symn & \icn & \symhash & \ichash & \symboth & \icboth & $D_O$ & \symn & \icn & \symhash & \ichash & \symboth & \icboth & $D_O$ & \symn & \icn & \symhash & \ichash & \symboth & \icboth \\
\midrule
English  & 98.0 & 96.4 & 96.0 & 91.5 & 90.4 & 90.8 & 90.5 & 95.2 & 93.4 & 88.4 & 68.9 & 65.0 & 68.7 & 64.4 & 90.3 & 80.5 & 78.5 & 56.0 & 53.0 & 54.2 & 51.8 \\
        
Chinese  & 93.1 & 93.5 & 91.9 & 88.7 & 88.5 & 88.7 & 87.6 & 69.8 & 73.9 & 66.9 & 51.5 & 41.5 & 51.6 & 43.5 & 81.0 & 86.8 & 80.5 & 55.7 & 52.3 & 58.1 & 53.3 \\
        
French   & 91.5 & 90.4 & 90.4 & 86.4 & 85.3 & 84.9 & 83.2 & 75.0 & 73.9 & 64.8 & 47.6 & 41.2 & 49.5 & 41.7 & 84.7 & 68.1 & 58.7 & 51.5 & 39.8 & 49.7 & 39.9 \\
        
Japanese & 89.9 & 86.3 & 84.8 & 83.1 & 81.5 & 81.2 & 82.1 & 69.8 & 65.9 & 54.5 & 43.0 & 34.8 & 42.3 & 35.2 & 79.0 & 76.9 & 68.1 & 52.2 & 44.0 & 52.1 & 44.5 \\
        
Swahili  & 91.9 & 92.6 & 90.9 & 85.1 & 84.4 & 85.1 & 83.9 & 63.7 & 64.4 & 50.8 & 41.6 & 32.4 & 39.8 & 31.2 & 86.7 & 82.4 & 74.0 & 54.7 & 47.6 & 53.5 & 48.1 \\
        
Amharic  & 82.3 & 82.3 & 81.5 & 74.5 & 73.1 & 75.4 & 73.6 & 15.3 & 18.7 & 6.0  & 11.9 & 3.9  & 11.8 & 3.5  & 39.9 & 45.5 & 40.5 & 25.9 & 23.4 & 26.6 & 24.1 \\
        
Igbo     & 78.2 & 79.1 & 76.7 & 68.1 & 66.0 & 70.0 & 65.9 & 43.5 & 44.6 & 33.8 & 26.4 & 18.3 & 27.0 & 18.5 & 44.4 & 45.3 & 40.2 & 26.7 & 21.4 & 27.7 & 22.7 \\
        
Yoruba   & 77.4 & 78.6 & 73.7 & 67.9 & 66.0 & 67.5 & 63.2 & 19.0 & 22.3 & 12.4 & 14.0 & 8.6  & 12.3 & 7.7  & 33.1 & 32.8 & 26.2 & 18.6 & 15.7 & 20.2 & 15.6 \\
        
Twi      & 52.8 & 52.9 & 46.0 & 43.2 & 36.2 & 44.8 & 37.5 & 14.9 & 14.1 & 8.3  & 8.5  & 5.0  & 9.5  & 3.4  & 18.5 & 19.1 & 16.5 & 10.4 & 9.2  & 12.3 & 9.8  \\

\midrule
Average  & 83.9 & 83.6 & 81.3 & 76.5 & 74.6 & 76.5 & 74.2 & 51.8 & 52.3 & 42.9 & 34.8 & 27.9 & 34.7 & 27.7 & 62.0 & 59.7 & 53.7 & 39.1 & 34.0 & 39.4 & 34.4 \\
\bottomrule
\end{tabular}%
}
\caption{Different models' accuracy across different dataset variations ($D_O$, \symn, \icn, \symhash, \ichash, \symboth, \icboth) for each language}
\label{tab:main_result_five}
\end{table*}

\begin{table*}[t] 
\centering
\resizebox{\textwidth}{!}{
\setlength{\tabcolsep}{3pt} 
\renewcommand{\arraystretch}{1.1}
\begin{tabular}{l ccccccc | ccccccc | ccccccc}
\toprule
& \multicolumn{7}{c}{\textbf{Qwen 3 32B}} & \multicolumn{7}{c}{\textbf{Qwen 3.5 27B}} & \multicolumn{7}{c}{\textbf{Gemma 2 9B}} \\
\cmidrule(lr){2-8} \cmidrule(lr){9-15} \cmidrule(lr){16-22} 
\textbf{Language} & $D_O$ & \symn & \icn & \symhash & \ichash & \symboth & \icboth & $D_O$ & \symn & \icn & \symhash & \ichash & \symboth & \icboth & $D_O$ & \symn & \icn & \symhash & \ichash & \symboth & \icboth \\
\midrule
English  & 89.1 & 84.7 & 87.3 & 87.7 & 86.5 & 86.0 & 86.0 & 98.4 & 97.2 & 94.2 & 93.8 & 90.7 & 92.0 & 88.9 & 75.9 & 77.9 & 76.7 & 45.3 & 47.3 & 44.7 & 48.5 \\
Chinese  & 89.9 & 90.9 & 88.8 & 87.8 & 88.1 & 88.2 & 86.1 & 89.9 & 91.5 & 90.1 & 87.3 & 83.9 & 86.4 & 85.1 & 78.4 & 83.3 & 76.8 & 49.0 & 46.1 & 49.2 & 44.2 \\
French   & 89.5 & 86.9 & 87.6 & 86.3 & 84.0 & 85.1 & 83.5 & 91.1 & 89.8 & 87.2 & 87.2 & 84.1 & 86.9 & 83.6 & 79.6 & 80.1 & 75.2 & 48.6 & 46.1 & 48.5 & 45.6 \\
Japanese & 88.7 & 86.2 & 85.2 & 84.8 & 83.3 & 84.3 & 82.9 & 87.5 & 83.7 & 80.1 & 76.0 & 72.1 & 76.1 & 68.5 & 75.1 & 70.0 & 64.4 & 45.1 & 39.3 & 43.2 & 39.9 \\
Swahili  & 78.2 & 79.2 & 71.0 & 73.6 & 67.3 & 71.3 & 64.6 & 91.9 & 90.9 & 88.5 & 88.1 & 84.9 & 88.2 & 84.0 & 69.8 & 71.3 & 68.6 & 44.4 & 42.4 & 43.2 & 43.8 \\
Amharic  & 42.7 & 45.8 & 39.4 & 37.7 & 33.6 & 38.6 & 30.6 & 77.0 & 79.4 & 76.0 & 68.1 & 67.4 & 67.3 & 64.0 & 41.6 & 43.5 & 36.8 & 24.2 & 20.7 & 24.2 & 18.9 \\
Igbo     & 19.8 & 20.2 & 14.9 & 14.7 & 10.0 & 15.0 & 10.8 & 73.8 & 74.7 & 70.8 & 69.5 & 62.9 & 70.8 & 63.2 & 31.6 & 34.0 & 29.5 & 17.1 & 17.1 & 18.1 & 16.6 \\
Yoruba   & 26.6 & 26.5 & 17.3 & 23.1 & 14.0 & 21.2 & 13.9 & 74.6 & 75.6 & 70.0 & 69.3 & 63.2 & 68.2 & 62.1 & 16.0 & 20.4 & 16.3 & 10.3 & 9.2  & 11.4 & 9.5  \\
Twi      & 5.6  & 5.7  & 3.5  & 4.6  & 2.2  & 5.6  & 2.2  & 37.5 & 35.9 & 27.4 & 31.3 & 25.4 & 32.4 & 25.5 & 8.0  & 10.5 & 7.8  & 6.7  & 3.3  & 7.0  & 3.1  \\

\midrule
Average  & 58.9 & 58.4 & 55.0 & 55.6 & 52.1 & 55.0 & 51.2 & 80.2 & 79.9 & 76.0 & 74.5 & 70.5 & 74.3 & 69.4 & 52.9 & 54.5 & 50.2 & 32.3 & 30.2 & 32.2 & 30.0 \\
\bottomrule
\end{tabular}%
}
\caption{Different models' accuracy across different dataset variations ($D_O$, \symn, \icn, \symhash, \ichash, \symboth, \icboth) for each language}
\label{tab:main_result_six}
\end{table*}

\end{document}